\pdfoutput=1
\documentclass[runningheads]{article}

\usepackage{mathtools}

\PassOptionsToPackage{numbers}{natbib}
\usepackage[final]{neurips_2022}

\usepackage[utf8]{inputenc} 
\usepackage[T1]{fontenc}    
\usepackage{hyperref}       
\usepackage{url}            
\usepackage{booktabs}       
\usepackage{amsfonts}       
\usepackage{nicefrac}       
\usepackage{microtype}      
\usepackage{xcolor}         
\usepackage{booktabs}
\usepackage{multirow}
\usepackage{overpic}
\usepackage{wrapfig}
\usepackage{caption}
\usepackage{subcaption}
\usepackage{soul}
\usepackage{authblk}
\usepackage{float}
\hypersetup{
    colorlinks=true,
    linkcolor=blue,
    filecolor=magenta,      
    urlcolor=magenta,
    pdftitle={Overleaf Example},
    pdfpagemode=FullScreen,
}

\usepackage{scalerel,stackengine,amsmath}

\title{Reduced Representation of Deformation Fields for Effective Non-rigid Shape Matching}
\newcommand{\etal}{\textit{et al.}}

\author[1]{Ramana Sundararaman}
\author[2,3]{Riccardo Marin}
\author[3]{Emanuele Rodol{\`a}}
\author[1]{Maks Ovsjanikov}

\affil[1]{LIX, Ecole Polytechnique, IP Paris}
\affil[2]{University of T{\"u}bingen}
\affil[3]{Sapienza University of Rome}

\begin{document}

\maketitle

\begin{abstract}
 In this work we present a novel approach for computing correspondences between non-rigid objects, by exploiting a reduced representation of deformation fields. Different from existing works that represent deformation fields by training a general-purpose neural network, we advocate for an approximation based on mesh-free methods. By letting the network learn deformation parameters at a sparse set of positions in space (nodes), we reconstruct the continuous deformation field in a closed-form with guaranteed smoothness. With this reduction in degrees of freedom, we show significant improvement in terms of data-efficiency thus enabling limited supervision. Furthermore, our approximation provides direct access to first-order derivatives of deformation fields, which facilitates enforcing desirable regularization effectively. Our resulting model has high expressive power and is able to capture complex deformations. We illustrate its effectiveness through state-of-the-art results across multiple deformable shape matching benchmarks. Our code and data are publicly available at: \url{https://github.com/Sentient07/DeformationBasis}.
 
 
\end{abstract}

\section{Introduction}
Shape correspondence is a central problem in computer vision and computer graphics as it facilitates many downstream tasks, such as tracking \cite{innmann2016volumedeform}, texture transfer \cite{boscaini2016learning} and statistical modeling \cite{bogo2014faust} to name a few. Due to its ubiquitous applicability, a wide range of techniques have been developed over the past several years \cite{sahilliouglu2020recent}. While early approaches relied on axiomatic modeling, recent methods follow data-driven techniques based on different input signals \cite{groueix20183d,litany2017deep,donati2020deep,zeng2020corrnet3d} within a shape collection.

A key question in this context is the choice of \emph{representation} used to model the non-rigid shape matching problem. Approaches based on \emph{intrinsic} or pose invariant representations have established a gold standard in the context where surfaces are well-defined \cite{ovsjanikov2012functional,eisenberger2020deep,Sharp2022}. Such methods, however, strongly rely on the presence of clean shapes and struggle when acquisition comes from noisy and non-uniform discretization \cite{Marin2020CorrespondenceLV}. In contrast, extrinsic techniques which directly operate on Euclidean space ($\mathbb{R}^3$) show strong resilience to artifacts. 

Unfortunately, this robustness of extrinsic methods often comes at the cost of relying on significant amounts of annotated training data \cite{donati2020deep,groueix20183d}. The main limiting factor arises in the representation of the deformation fields. The standard approach is to use general-purpose MLPs to learn deformation fields that can fit an arbitrary shape deformation~\cite{deng2021deformed,zheng2021deep,groueix20183d}. However, given the fact that MLPs are general-purpose networks, they require significant amounts of training data to learn both coarse and fine details \cite{groueix20183d}. 

\begin{figure}[!t]
	\centering
	\footnotesize
	\includegraphics[width=\textwidth]{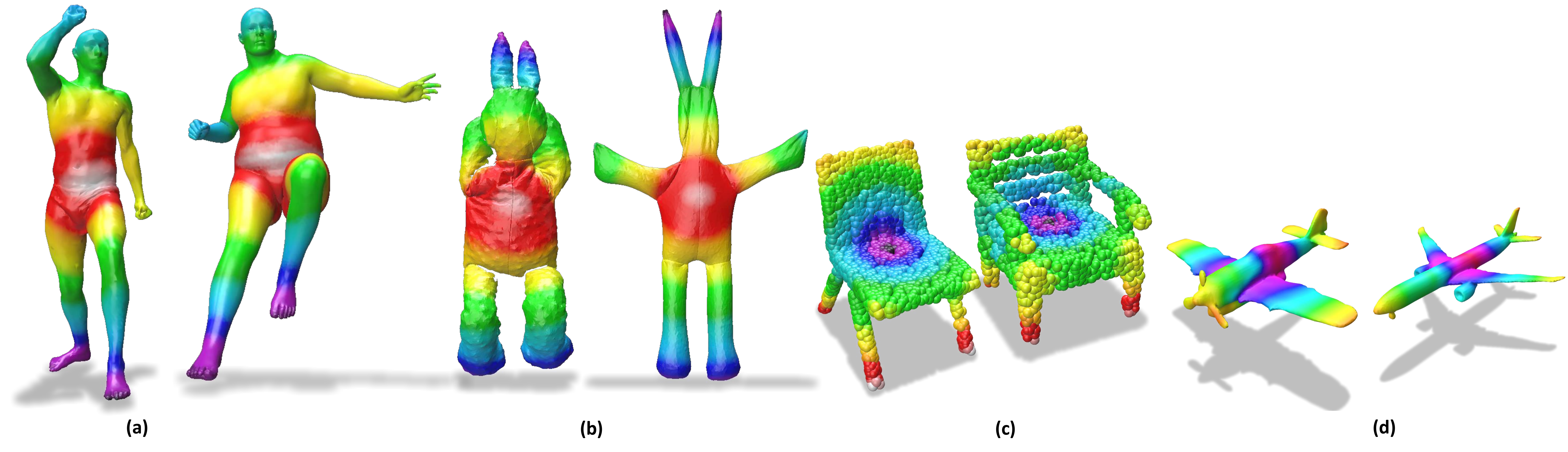}
	\caption{\label{fig:teaser} Examples showing generalization of our deformation field representation. Our approach allows to compute correspondences across a wide range of shape categories: (a) human articulation, (b) physically-based deformation from real-world scans, (c) shapes undergoing topological changes, and (d) shapes represented via implicit functions. \vspace{-2mm}}
\end{figure}

To overcome this limitation, we propose to learn a coarse representation of \emph{deformation parameters} at fixed positions in space called ``nodes''. By learning a reduced representation of deformation fields, intuitively, we restrict the learning process to global patterns of the input signal. Then, to recover finer details, we reconstruct the continuous deformation field function in closed-form using a class of mesh-free approximation techniques~\cite{Lancaster1981SurfacesGB}. This allows us to scale our approach to arbitrary resolution with guaranteed smoothness and across different object classes as shown in Figure~\ref{fig:teaser}.

Apart from being theoretically grounded and simple in practice, our reduced representation has two key advantages. First, it is significantly more data-efficient and can learn to capture complex deformations given only a small number of examples. Secondly, it is more amenable to regularization, since it provides explicit access to first-order derivatives of the deformation field in closed form. This is especially useful in imposing geometric priors such us local rigidity and volume preservation. 

Our contributions can be summarized as follows: (a) We propose to learn a compact representation of deformation parameters, that is data-efficient, resolution agnostic, and facilitates regularization through direct access to deformation gradients (b) We show an efficient way of incorporating desirable regularization to promote a well-structured deformation space. (c) Through extensive experiments across real-world and synthetic datasets, we demonstrate the generalization ability of our method over different down-stream applications such as non-rigid shape matching, registration, unsupervised part segmentation and interpolation.

\section{Related Works}
\subsection{Non-rigid shape Correspondence and Registration}

Shape correspondence is a very well-studied area of computer vision and computer graphics and we refer interested readers to the recent survey ~\cite{sahilliouglu2020recent} for a comprehensive overview. Notable axiomatic approaches in this category are based on the functional maps paradigm~\cite{ovsjanikov2012functional,kovnatsky2013coupled,aflalo2013spectral,rodola2017partial}, that aims to compute a near-isometric mapping by estimating a linear transformation between functions represented in a reduced basis. This framework has been successfully adapted by learning techniques~  \cite{litany2017deep,halimi2019unsupervised,roufosse2019unsupervised,donati2020deep,eisenberger2020deep} which demonstrate near-perfect accuracy~\cite{Sharp2022} on several shape correspondence benchmarks. However, these approaches can be prone to errors in the presence of noisy point clouds or significant acquisition artefacts. Although registration-based techniques~\cite{FARM,HirshbergECCV2012,TransMatch} present a relatively more robust option, they are often based on human-centric priors or require significant training data. 

\subsection{Template-based and Template-free Methods}
Deforming a template shape to match a target geometry is a long-standing and well studied problem~\cite{allen_articulated,allen_articulated_2,VolkerMorphable}. Such a template can be a polygonal mesh~\cite{Krishnamurthy1996FittingSS,MatthiasMultiRes}, possibly parameterised~\cite{SMPL2015,ZuffiCVPR2017,Hasler2009} or an unordered point-set~\cite{groueix2018b,Deprelle2019LearningES,groueix19cycleconsistentdeformation,Wang_2019_CVPR,groueix2018} or implicitly defined through zero-level set of a Neural Field~\cite{deng2021deformed,zheng2021deep,atzmon2021augmenting}. In the recent years, learning based model-free deformation techniques~\cite{groueix2018b,Deprelle2019LearningES,TransMatch} have emerged as a viable option for registration and correspondence tasks given copious amount of training data~\cite{groueix2018b,ZuffiCVPR2018}. Among them, the closest to our approach is 3D-CODED~\cite{groueix2018b} which learns deformation fields through point-wise MLPs. However, since this approach fits a general-purpose MLP and treats each point on the shape equally likely, it requires abundant training data to achieve optimal performance.

\subsection{Deformation Field Representation and Shape Interpolation}
Deformation between a pair of shapes can be represented as a simple displacement field at every vertex. However, such a representation can be unnecessarily complex, and costly to optimize. As a result, several alternatives have been proposed. The most prominent ways to parameterize the space of deformations include handle-based~\cite{Schaefer2006ImageDU,MaksSCA,DeepMLS,JacobsonBiHarmonic} or cage-based~\cite{LipmanCage,Floater2005,PushkarHarmonic,YifanNeuralCage2020} representations (see also \cite{bechmann1994space,gain2008survey,nieto2013cage} for an overview). More recently, a common approach is to construct a reduced representation via a learned latent embedding~\cite{uy-joint-cvpr21,jiang2020shapeflow,Cosmo2020,Rakotosaona2020}.

\paragraph{Deformation Field regularization} Several geometric constraints have been proposed with the aim of preserving desirable properties of the shape by the deformation field, including imposing elasticity~\cite{ARAP,GrinspunShell,Iglesias2017} and volume preservation~\cite{RappoportVolume,ZhouVolume,MaksSCA,Hirota2000,Eisenberger2019DivergenceFreeSC}. Recently, these constraints have been successfully adapted by data-driven methods~\cite{neuromorph,jiang2020shapeflow,park2021nerfies,atzmon2021augmenting,yang2021geometry} and more relevantly through the differential of the map~\cite{jiang2020shapeflow,yang2021geometry,yang2021geometry}. Distinct from such approaches, our approximation via mesh-free method enables evaluating this map differential at fixed points in a closed-form, which significantly simplifies the deformation field regularization without additional computational overhead.

\paragraph{Shape Interpolation} Shape interpolation refers to time-parameterized deformation, where a source shape is continuously deformed to a target shape. Our work is related to efforts which aim to enforce intermediate shapes to preserve certain intrinsic properties~\cite{Cosmo2020,Rakotosaona2020,neuromorph,Heeren2018,Tan2020,jiang2020shapeflow}. Among them closest to our approach is LIMP~\cite{Cosmo2020} which disentangles the latent space based on style and pose to preserve geodesic distance. In contrast, our approach does not require such a priori information, which can be costly in terms of annotation efforts. 

\subsection{Reduced representations and Approximations}

In this work we use mesh-free function approximation method~\cite{Belytschko1996,Fries2004ClassificationAO,Lancaster1981SurfacesGB}, to approximate deformation fields. Mesh-free methods have been successfully adapted in Smoothed Particle Hydrodynamics (SPH) modeling~\cite{Hammani2020,Marrone2011}, image processing~\cite{Schaefer2006ImageDU}, animation~\cite{MaksSCA,Mller2004,Mller2005} and more recently in a data-driven framework~\cite{DeepMLS}. Differently from ~\cite{DeepMLS}, instead of learning the weights of the least squares function, we instead learn deformation values at nodes and demonstrate our method to be applicable in wide-range of downstream tasks. Alternatively, Eisenberger~\etal~\cite{Eisenberger2019DivergenceFreeSC} have proposed to use a compact representation of deformation fields using the first $k$ eigenfunctions of the Laplace Beltrami Operator (LBO). While their approach provides volume preserving deformation, it does not facilitate other regularizations such as as-rigid-as-possible deformation fields without requiring correspondence at inference time~\cite{Eisenberger2020HamiltonianDF}.

\section{Motivation, Background and Notation}
\label{sec:motivation}

\subsection{Motivation:}

Parametric models such as SMPL~\cite{SMPL2015} have been tremendously useful over the recent years in digitizing and processing human models. This success can largely be attributed to their expressive power, allowing to generate a wide range of styles and poses using a small fixed set of deformation parameters. While this efficacy with such a compact representation is remarkable, it also raises an inspiring question: what is the optimal amount of \emph{learnable} parameters necessary to represent general deformations? Today, general-purpose MLPs form the conventional way of representing deformation fields due to their simplicity and potential of being universal functional approximators~\cite{Hornik1989,Pinkus1999}. 
Unfortunately, the generic power of MLPs also comes at a cost of copious training efforts~\cite{groueix2018b,TransMatch}. Furthermore, representing a deformation field using a neural network makes access to certain quantities such Jacobian matrices of deformation fields cumbersome. For these reason, we propose to learn a reduced set of deformation parameters from which we \emph{approximate} the deformation field function using a mesh-free method.

\subsection{Mesh-free Approximation}

Mesh-free methods are a class of approximation techniques which constructs a continuous function based on independent, potentially sparse and irregular observations. Assume that our domain of interest $\mathbb{R}^{3}$ is equipped with $K$ fixed points $\mathbf{q}_i \in \mathbb{R}^3$ along with some observations $u_i$ at $\mathbf{q}_i$ and a choice of a polynomial basis $p(\cdot)$. We refer to fixed points $\mathbf{q}_i$ as ``nodes'' (or, alternatively, ``deformation nodes''). Our main goal is to construct a continuous approximation of some real-valued function $u(.)$ in some subdomain $\Omega \subset \mathbb{R}^{3}$ of interest. We let $\mathbf{x} \in \Omega$ to be an arbitrary point in our region of interest. The key idea behind this approximation is to use a \emph{local weighted least-squares} fitting (also referred to as Moving Least Squares)  approach~\cite{Lancaster1981SurfacesGB}. Specifically, we first build a compactly supported weighting function $w_i(\mathbf{x})$ in the neighborhood of $\mathbf{q}_i$, via:
\begin{equation}
   w_i(\mathbf{x})= 
\begin{dcases}
    \left(1 - \frac{ ||\mathbf{x} - \mathbf{q}_i||^{2}_{2}} {r_{i}}\right)^{3},& \text{if } ||\mathbf{x} - \mathbf{q}_i||^{2}_{2}\leq r_i\\
    0,              & \text{otherwise}
\end{dcases}
\end{equation}

The compactness of this weighting function is useful in preserving the local characteristics of approximation. From this, a \emph{Shape Function} $\Phi_i$ associated with each node $i$, is constructed as:
\begin{equation}
\Phi_i(\mathbf{x})=p^{T}(\mathbf{x})[M(\mathbf{x})]^{-1} w_i(\mathbf{x}) p\left(\mathbf{q}_i\right).
\label{eqn:Phi}
\end{equation}

Here $M(\mathbf{x})$ is the Moment Matrix associated with the approximation, and defined as:
$$M(\mathbf{x})=\sum_{i=1}^{K} w_i(\mathbf{x}) p\left(\mathbf{q}_i\right) p^{T}\left(\mathbf{q}_i\right)$$

The shape function $\Phi_i$ is a continuous function that describes how each node $\mathbf{q}_i$ influences the approximation of $u(.)$ across points $\mathbf{x} \in \Omega$. Jointly the the set of $\Phi_i$'s enable the reconstruction of arbitrary functions up to $n^{th}$ order consistency~\cite{Fries2004ClassificationAO}, where, $n$ is the order of the polynomial $p(\cdot)$. Specifically, a smooth local approximation of $u(\mathbf{x})$ is given as:
\begin{equation}
u(\mathbf{x})=\sum_{i=1}^{K} \Phi_{i}(\mathbf{x}) u_{i}
\label{eqn:FieldApprox}
\end{equation}

 As the construction of $\Phi$ involves computing $M^{-1}$ (c.f Eq.~\eqref{eqn:Phi}), it is a sufficient condition for each point $\mathbf{x}$ to be compactly supported by 4 non-planar nodes $\mathbf{q}_i$ for $M$ to be non-singular. It is important to note that Eq.~\eqref{eqn:FieldApprox}  is approximating and \emph{not interpolating}, i.e $u_i \neq u(\mathbf{q}_i)$. For instance, owing to the compact nature of $w(\mathbf{x})$, it is possible that $u(\mathbf{q}_i)$ is undefined if $\mathbf{q}_i \notin \Omega$. For this reason, we sample the nodes a priori to have a well-supported domain $\Omega \subseteq \mathbb{R}^3$ where $u(\mathbf{x})$ is well-defined. 
 
 Furthermore, an important advantage of using mesh-free approximations comes from an exact analytical expression for the gradient function of $u(\mathbf{x})$. To the scope of our current discussion, considering $u(\mathbf{x})$ to be the approximation of deformation field function, the Jacobian of this deformation field only depends on evaluation point and is independent of observed deformation parameters $u_i$ ,

\begin{equation}
    \mathbb{J} = \nabla_{x,y,z}u(\mathbf{x}) = \sum_{i=1}^{K} \left[\frac{\partial \Phi_{i}(\mathbf{x})}{\partial x}, \frac{\partial \Phi_{i}(\mathbf{x})}{\partial y}, \frac{\partial \Phi_{i}(\mathbf{x})}{\partial z}\right]^{T} u_{i}
\label{eqn:anlyGrad}
\end{equation}

This nice property helps us characterize the deformation fields with desired first-order regularization in an efficient manner. We refer interested readers to~\cite{Fries2004ClassificationAO,Belytschko1996} for a detailed summary.

\subsection{Notation: } 
As our training set, we consider a collection of shapes $\{\mathcal{S}_{1} \ldots \mathcal{S}_{N}\}$ with ground truth correspondences $\Pi_{\mathcal{S}_{l}\mathcal{S}_{j}}$ between them. Shapes can be represented as triangular meshes $\mathcal{S}_{j} \coloneqq \{\mathcal{V}, \mathcal{E}\}$ or simply unordered sets of points (point clouds) $\mathcal{S}_{j} \coloneqq \{\mathcal{V}\}$. We pick one shape from the collection as a template $\mathcal{T}$, and let $[\mathcal{T}]$ be the volume enclosed by the boundary $\partial \mathcal{T}$. We refer to $\mathcal{Q} \in \mathbb{R}^{K \times 3}$ as \emph{nodal positions}, which are $K$ \emph{fixed} points in space sampled from the template volume $[\mathcal{T}]$.  We let $\mathcal{D} (\cdot): \mathbb{R}^{3} \rightarrow \mathbb{R}^{3}$ be the \emph{deformation mapping}, which, intuitively maps points in the deformation volume to points on target shapes. We refer to $U_j$ as the nodal deformation parameters corresponding to the $j^{\text{th}}$ shape and analogously define $D_j(\cdot)$. Each node $q_{i} \in \mathcal{Q}$ has a support radius $r_{i}$ and associated deformation parameter $u_{i}$. We use lower-case notation $u_{i,j}$ to refer to the value of  the deformation field at node $\mathbf{q}_i$ corresponding to shape $j$. We denote $\mathbf{x} \in \Omega \subset \mathbb{R}^3$ as points in space which are supported by at least four non-planar nodes. We refer to $U_j(\mathbf{x})$ as the continuous approximation of the deformation field, constructed from deformation parameters $U_j$ using Eq.~\eqref{eqn:FieldApprox}. We re-iterate that $u_{i,j} \neq U_j(\mathcal{Q}_i)$. The relation between a deformation \textit{field} and a deformation \textit{mapping} is given by $\mathcal{D}_j (\mathbf{x}) \coloneqq \mathbf{x} + U_j(\mathbf{x})$. For the sake of consistency, we index nodes using $i$, shape collection using $j,l$ and points within shape using $k$.

\section{Method: Learning Nodal Deformation-Field}

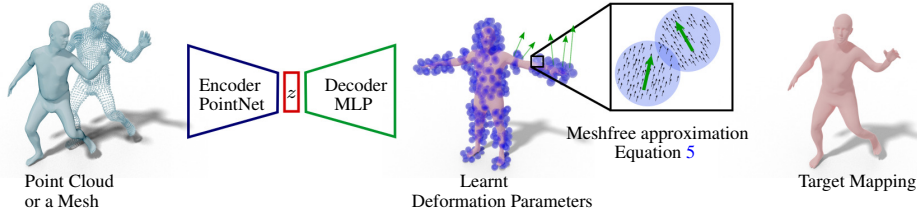
\begin{figure}[!t]
	\centering
	\footnotesize
	
	\begin{overpic}[trim=0cm 0cm 0cm 0cm,clip, width=0.92\linewidth]{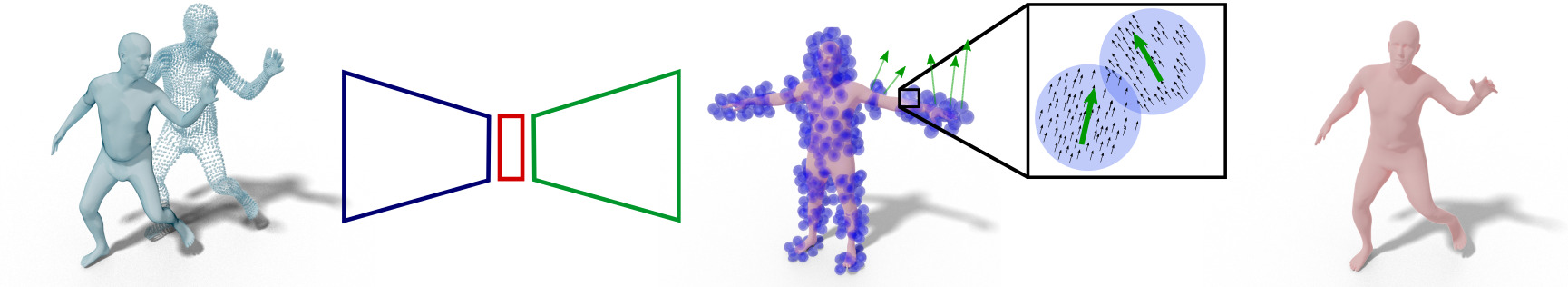}
	\scriptsize
		\put(23,9){{Encoder}}
		\put(23,7){PointNet}
	
	   \put(32.0,8.3){$z$}
		
		\put(36,9){Decoder}
		\put(37,7){MLP}
		
		\put(5,-1){Point Cloud}
		\put(5,-3){or a Mesh}
		\put(50,-1){Learnt}
		\put(45,-3){Deformation Parameters}
		
		\put(61,4){Meshfree approximation}
		\put(66,2){Equation \ref{eqn:LSApprox}}
		
		\put(85,-1){Target Mapping}
	\end{overpic}
	\vspace{0.4cm}
	\caption{\label{fig:architecture} Overview of our approach. First, we learn the deformation parameters at nodes using an Auto-Encoder. Then, we use mesh-free approximation to obtain a continuous deformation mapping.}

\end{figure}

\paragraph{Overview.}
Our network is based on a PointNet~\cite{qi2016pointnet} auto-encoder as shown in Figure~\ref{fig:architecture}. Our network $\mathcal{F}_{\theta} (\cdot)$ predicts nodal deformation parameters $U_j$ for each training shape $\mathcal{S}_{j}$, i.e $U_j = \mathcal{F}_\theta(\mathcal{S}_j)$ where $U_j \in \mathbb{R}^{K\times3}$. As mentioned before, the nodes $\mathcal{Q}$ are fixed a priori. From the predicted $U_j$, we can compute the shape-specific deformation mapping $\mathcal{D}_{j}(x)$ and its Jacobian $\mathbb{J}_j$, via:

\begin{equation}
\begin{split}
\mathcal{D}_{j}(\mathbf{x}) &=\mathbf{x} + \sum_{i=1}^{K} \Phi_{i}(\mathbf{x}) u_{i,j} \\
\mathbb{J}_j &= \mathbf{I} + \nabla_{x,y,z}U_j(\mathbf{x})
\end{split}
\label{eqn:LSApprox}
\end{equation}

Where, $\nabla_{x,y,z}U_j(\mathbf{x})$ is given in Equation~\ref{eqn:anlyGrad}.

\subsection{Training}
\label{subsec:training}
Intuitively, we would like to train a network so that $S_j \approx \{ D_{j}(\mathbf{x}) | \mathbf{x} \in \mathcal{T} \}$, subject to appropriate regularization. 
Although $\mathcal{D}_{j}(\mathbf{x})$ can be approximated at an arbitrary $\mathbf{x}$, which is supported by four non-planar nodes, we restrict ourselves to $\mathbf{x} \in \mathcal{T}$ for the ease of learning. As $\mathcal{F}_{\theta} (\cdot)$ represents an auto-encoder, it can be decomposed as $\mathcal{F}_{\theta} (\mathcal{S}_{j}) = Dec(Enc(\mathcal{S}_{j})) = Dec(Z_{j})$ where $Z_{j}$ denotes the latent embedding. Leveraging this fact, we provide a novel way to promote plausible latent deformation spaces $Z_{j}$ by enforcing first-order constraints over the \emph{intermediate shapes} as well. The overall optimization objective of our network is given as:
\begin{equation}
    \mathcal{L}_{net} = \lambda_{1} \mathcal{L}_{cor} + \lambda_{2} \mathcal{L}_{vol} + \lambda_{3} \mathcal{L}_{arap} + \lambda_{4} \mathcal{L}_{Z}
\label{eqn:optimObj}
\end{equation}

For the unsupervised case, we replace $\mathcal{L}_{cor}$ with $\mathcal{L}_{CD}$ which denotes the Chamfer's distance.
\paragraph{Correspondence Loss}

Given a set of $\mathcal{C}$ of corresponding points $\{\mathbf{x}_{l}, \mathbf{x}_{k}\}$, where $\mathbf{x}_{l} \in \mathcal{T}$, $\mathbf{x}_{k} \in \mathcal{S}_j$ our correspondence loss is given as 

\begin{equation}
\mathcal{L}_{cor}=\sum_{j=1}^{N}\sum_{(\mathbf{x}_k, \mathbf{x}_l) }^{|\mathcal{C}|}\left\|\mathcal{D}_{j}\left(\mathbf{x}_{l}\right)-\mathbf{x}_{k}^{j}\right\|^{2}_{2}
\end{equation}

Where $x_{k}^{j}$ denotes the $k^{th}$ point in $j^{th}$ shape. 
\paragraph{Volume Preserving Field}
A deformation field is volume preserving iff its Jacobian has unit determinant over the entire shape. Consequently, our local volume preservation regularization is given as follows:

\begin{equation}
\mathcal{L}_{vol}=\sum_{j=1}^{N}\sum_{i=1}^{K} |\operatorname{det}(\mathbb{J}_{j}(\mathbf{q}_{i}))-1|^{2}_{2}
\label{eqn:volP}
\end{equation}

We empirically observe poor convergence when this objective is enforced over the entire shape due to its stringent nature. This is because not all deformations are strictly volume preserving. Thus, we restrict this regularization only at nodes.

\paragraph{As Rigid As Possible (ARAP) Deformation}

Since rigid motions preserve pairwise distances, a deformation field associated with such a transformation is characterized by an orthonormal Jacobian matrix. Thus, in order to promote locally rigid deformation field at the deformation nodes, we define our ARAP regularization as:

\begin{equation}
\mathcal{L}_{arap} = \sum_{j=1}^{N}\sum_{i=1}^{K} \left\| \mathbb{J}_{j}^{T}\left(\mathbf{q}_{i}\right)  \mathbb{J}_{j}\left(\mathbf{q}_{i}\right)-\mathbf{I}\right\|_{F}^{2}
\label{eqn:arap}
\end{equation}

\paragraph{Structuring Latent Deformation Space}
A well-known advantage of an auto-encoder architecture is the construction of the latent space, where each shape has an embedding $Z_{j} \in \mathbb{R}^{D}$. Then, a parameterized path in this latent space between two shapes $Z_{l,j} (\alpha) = \alpha Z_{l} + (1-\alpha)  Z_{j}$ continuously deforms $\mathcal{S}_{j}$ to $\mathcal{S}_{l}$ with rate of change controlled by $\alpha$. This allows constructing a sequence of shapes, often referred to as interpolated shapes. Since each $Dec(Z_{j}) = U_{j}$, we can further require our network $\mathcal{F}_\theta$ to produce a plausible deformation between \emph{each pair of training shapes}. To that end, we introduce our latent smoothness loss as follows:

\begin{equation}
\mathcal{L}_{Z} = \sum_{l \neq j}^{|\mathcal{S}|} \mathcal{L}_{arap}\left(\operatorname{dec}((1-\alpha) \mathbf{z}_{j}+\alpha \mathbf{z}_{l})\right)  + \sum_{l \neq j}^{|\mathcal{S}|} \mathcal{{L}}_{vol}\left(\operatorname{dec}((1-\alpha) \mathbf{z}_{j}+\alpha \mathbf{z}_{l})\right)
\label{eqn:lossZ}
\end{equation}

Where $\mathcal{L}_{arap}, \mathcal{L}_{vol}$ are defined in Eqs.~\eqref{eqn:volP},~\eqref{eqn:arap} and $\alpha \in (0,1)$ are sampled randomly.

\subsection{Inference}
At test-time, given a pair of unseen shapes $(\mathcal{X}, \mathcal{Y})$ we follow a three-step procedure to obtain the correspondence $\Pi_{\mathcal{X}\mathcal{Y}}$. First, we separately reconstruct $(\mathcal{D}_\mathcal{X}, \mathcal{D}_\mathcal{Y})$ by deforming the fixed template $\mathcal{T}$. Second, we enhance the respective reconstructions by optimizing the latent vector $Z$ independently for shapes $(\mathcal{X}, \mathcal{Y})$. The objective for this optimization is to minimize the bi-directional Chamfer Distance~\cite{groueix2018b} while \emph{also} enforcing first-order constraints as follows:
\begin{equation}
Z=\underset{Z}{\operatorname{argmin}} \hspace{1mm} \Lambda_{1} \mathcal{L}_{\mathcal{CD}}+\Lambda_{2} \mathcal{L}_{arap}+\Lambda_{3} \mathcal{L}_{vol}.
\label{eqn:Refinement}
\end{equation}

As $\mathcal{D}_\mathcal{X}$ is the reconstruction of $\mathcal{X}$, the correspondence between $\mathcal{D}_\mathcal{X}, \mathcal{X}$ can be computed via a simple nearest neighbor search in 3D (analogously for $\mathcal{Y}$). Since $\mathcal{D}_\mathcal{X}, \mathcal{D}_\mathcal{Y}$ are deformed versions of a template they enjoy a natural correspondence (by vertex ordering). Finally, the correspondence between $(\mathcal{X}, \mathcal{Y})$ is a composition of two nearest neighbour searches $\Pi_{\mathcal{X}\mathcal{Y}} == (\text{NN}(\mathcal{D}_\mathcal{X}, \mathcal{X}), \text{NN}(\mathcal{D}_\mathcal{Y}, \mathcal{Y}))$.

\subsection{Extending sparse to dense Correspondence}
An added advantage of our representation is the ability to retrieve dense shape correspondence between a shape pair, given a few sparse key-point correspondences $(\mathbf{x}_{l}, \mathbf{y}_{k}), \forall  \mathbf{x}_{l} \in \mathcal{X},\forall \mathbf{y}_{k} \in \mathcal{Y}$. First, we estimate the deformation parameter $u_{i}$ at the nodes by solving an optimization:

\begin{equation}
u_{i}=\underset{u_{i}}{\operatorname{argmin}} \hspace{1mm} \lambda_{1}\sum_{\forall (\mathbf{x}_{l}, \mathbf{y}_{k})}^{} \left\|\mathcal{D}_{\mathcal{X}}\left(\mathbf{x}_{l}\right)-\mathbf{y}_{k}\right\|^{2}_{2} + \lambda_{2} \left\|\mathbb{J}_{\mathcal{X}}^{T} \mathbb{J}_{\mathcal{X}}-I\right\|_{F}^{2} + \lambda_{3}  |\operatorname{det}(\mathbb{J}_{\mathcal{X}})-1|^{2}_{2}
\label{eqn:propogation}
\end{equation}

Then, a dense mapping can be computed by approximating the deformation field (c.f Equation~\ref{eqn:LSApprox}). 

\subsection{Implementation details}

\paragraph{Analytical Gradients and Timing advantages}
We leverage the advantage of inexpensive access to Jacobians as mentioned in Equation~\ref{eqn:anlyGrad}. Because our evaluation points are known a priori, due to the use of a fixed template $\mathcal{T}$, the matrix $\mathbb{J}$ can be pre-computed and re-used at training and evaluation. In practice, we observe a $\mathbf{10\times}$ speed-up at training time when enforcing our first-order constraints and a $\mathbf{350\times}$ speed-up incorporating the latent constraints (c.f. Eqn~\ref{eqn:lossZ}). We provide more timing details in the supplementary.

\paragraph{Node Sampling:}

Since the deformation field at a point is determined by the nodes within the radius, it is important to limit the influence of a node which is close in a Euclidean sense but geodesically far. For instance, it is counter-intuitive to have a node in the trunk of the human influencing the deformation of a point in the arm. Bearing this in mind, our node sampling strategy is divided into three main steps. First, we construct a dense sampling of points in the volume and around the boundary of the template $\partial \mathcal{T}$. Second, we use rejection sampling to \emph{exclude} a node that exerts its influence in semantically different regions~\cite{SMPL2015}. Finally, we perform Farthest Point Sampling (FPS) until each surface point is covered by 4 non-planar nodes. We emphasize that this step is performed only on the template shape and using SMPL~\cite{SMPL2015} segments is \emph{one of many} possible ways to perform segmentation. An in-depth ablation study is provided in the supplementary material.

\section{Experiments}
The reduced representation for deformation field which we have discussed so far is conducive to produce naturally smooth deformation while significantly reducing the amount of supervision needed to facilitate learning. We empirically show the efficacy of our proposed representation of deformation fields across four main tasks, namely Non-rigid 3D shape correspondence, Shape registration, Unsupervised part segmentation and Shape interpolation.

\begin{figure}[!t]
	\centering
	\footnotesize
	
	\begin{overpic}[trim=0cm 0cm 0cm 0cm,clip, width=0.9\textwidth]{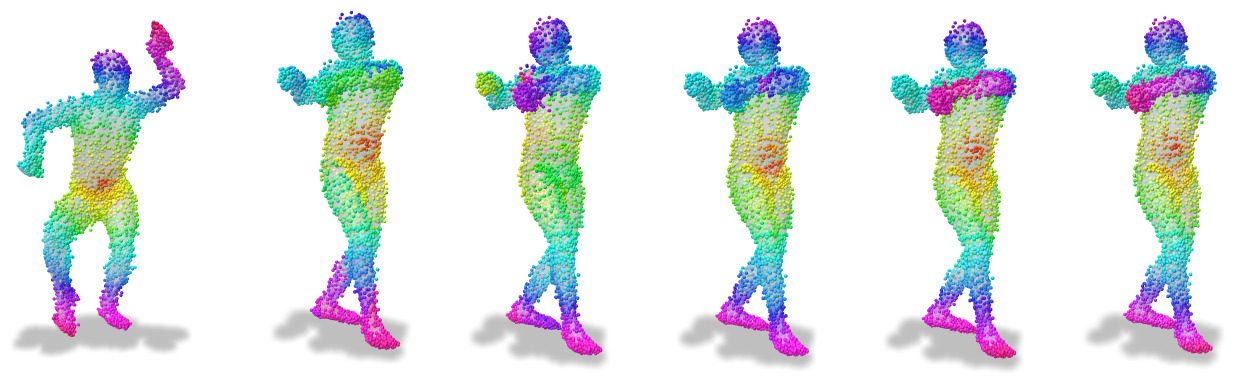}
		\put(5,30){Source}
		\put(26,30){\cite{Marin2020CorrespondenceLV}}
		\put(42,30){\cite{groueix2018b}}
		\put(59,30){\cite{TransMatch}}
		\put(75,30){Ours}
		\put(91,30){GT}
	\end{overpic}
	
	\caption{\label{fig:matching} Color-coded correspondences on the SCAPE (PC+N) dataset. ``Twist'' is a challenging articulation as a wrong deformation can lead to large geodesic error (see Cheese-Pull effect in \cite{Eisenberger2020SmoothSM}, Figure 11). We hypothesize that our approach, which learns a ``global'' sense of the articulation, does not suffer from such artefacts since the fine (local) details are computed in closed form.}

\end{figure}
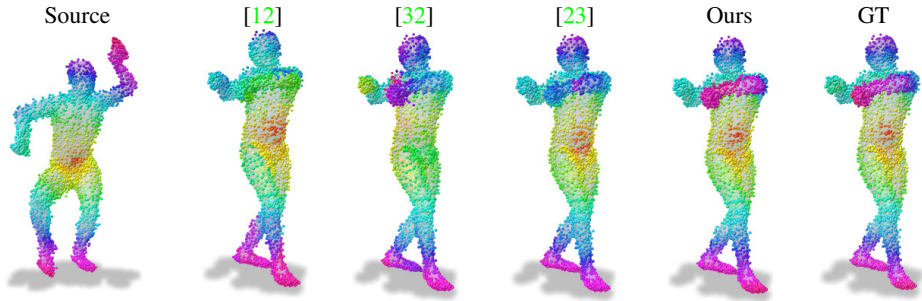

\begin{table}
\centering
\scriptsize 

\begin{tabular}{llr|rrr} 
\midrule
\multicolumn{3}{c}{Method}                                             & \multicolumn{3}{c}{Correspondence Error}  \\ 
\midrule
Type                            & Name       & \#Tr data               & SHREC’19 & FAUST(NI)~ & SCAPE(PC+N)~      \\ \midrule
\multirow{1}{*}{Spectral}    & GeoFMap~\cite{donati2020deep}    & 1.7 & 11.2     & 20.1       & 27.7              \\ \midrule
\multirow{2}{*}{Pair-wise}      & Diff-FMap~\cite{Marin2020CorrespondenceLV}  & 1.0                       & 15.1     & 5.4        & 26.0              \\
                                & CorrNet3D~\cite{zeng2020corrnet3d}  & 15.0                      & 9.6      & 25.9       & 38.0              \\ 
\midrule
\multirow{3}{*}{Template based} & 3D-CODED~\cite{groueix20183d}    & 23.0                      & 10.3     & 7.0        & 18.7              \\
                                & TransMatch~\cite{TransMatch} & 1.0                       & 6.1      & 6.5        & 17.1              \\
                                & Ours       & \textbf{0.1}                     & \textbf{4.8}      & \textbf{5.3}        & \textbf{6.6}               \\
                                \midrule
\end{tabular}
\caption{We report correspondence error as geodesic distortion (in cm) scaled by square root of shape area. \#Tr data denotes number of training shapes scaled by $10^{-4}$. }
\label{tab:correpResult}
\end{table}

\subsection{Shape Correspondence}
\label{subsec:NRSC}

We consider three challenging benchmarks, namely, SHREC'19, FAUST (PC), SCAPE (PC+N). SHREC'19~\cite{SHREC19} is a standardised benchmark consisting of 430 evaluation pairs with significant variations in mesh resolution and connectivity. FAUST (PC) denotes a more recent Non-Isometric Point Cloud variant~\cite{Marin2020CorrespondenceLV} of the FAUST dataset consisting of 1000 points with large variance in point sampling density. Third, we evaluate on a variant of the recent SCAPE-Remesh dataset~\cite{Ren2018} consisting of 20 shapes of the same human in 20 distinct poses. We further augment the challenge by adding random Gaussian noise and refer to as SCAPE (PC+N). We evaluate correspondence error following the Princeton benchmark protocol~\cite{KimBIM}. We train our method on a subset of 1000 SURREAL shapes~\cite{varol17_surreal} for 1000 epochs with data-augmentation along Y-axis.

\paragraph{Baselines}
We compare our method against data-driven correspondence methods broadly classified into Spectral, Pairwise and Template based. We use GeoFMap~\cite{donati2020deep} with the more robust feature extractor Diffusion-Net~\cite{Sharp2022} as our spectral baseline, Diff-FMaps~\cite{Marin2020CorrespondenceLV} and CorrNet3D~\cite{zeng2020corrnet3d} as our pair-wise baseline. For our template based baselines, we use 3D-CODED~\cite{groueix2018b} and TransMatch~\cite{groueix2018b}. For the evaluation of baselines on our proposed SCAPE (PC+N), we use the author-provided pre-trained models, and apply consistent pre-processing to the input shapes across all methods. 
\paragraph{Discussion}
Our approach consistently outperforms baselines as summarised in Table~\ref{tab:correpResult}. While our quantitative correspondence results are persuasive, it is remarkable to note that our method requires an order of magnitude less training data in comparison to competing methods. This supports our premise that characterizing typical deformations requires far fewer parameters than what is leveraged by existing data-driven methods. We show qualitative correspondence results through color transfer for a challenging pair with ``twisted'' motion in Figure~\ref{fig:matching}.

\subsection{Shape Registration}

Shape registration is a special case of correspondence, where our goal is to find an optimal deformation between the scan and a fixed template. For this, we consider the recent SHREC'20 benchmark~\cite{DykeShrec20}, consisting of 11 partial scans of stuffed toy rabbits to be registered to a single scan. This benchmark is particularly challenging due to granulated surface deformation, scanning artefacts, and limited data and supervision.

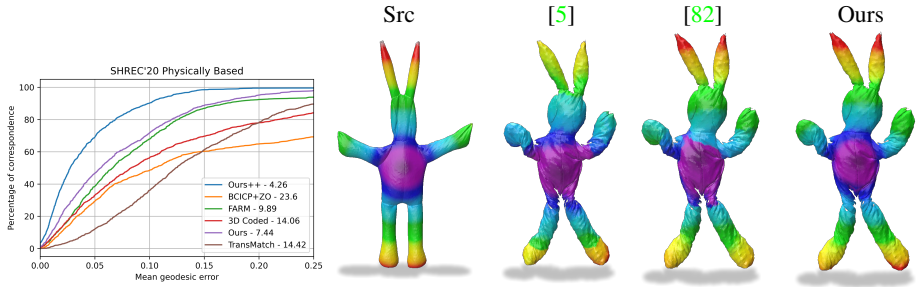
\begin{figure}[!t]
	\centering
	\footnotesize
	
	\begin{overpic}[trim=0cm 0cm 0cm 0cm,clip, width=0.92\linewidth]{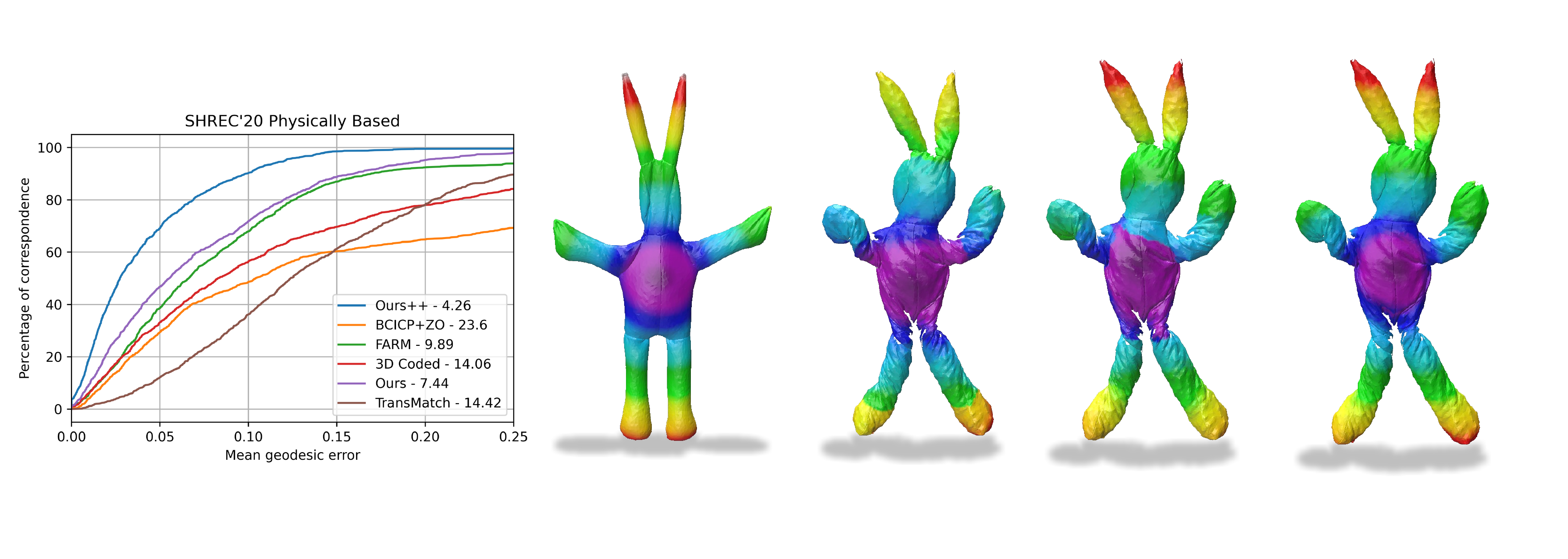}
		\put(40,33){Src}
		\put(57,33){\cite{groueix20183d}}
		\put(71,33){\cite{marin2020farm}}
		\put(87, 33){Ours}
	\end{overpic}

	\caption{Quantitative and qualitative results on SHREC'20. Our approach predicts smooth correspondences across highly-granular surface-level deformation.}
    \label{fig:registration}
\end{figure}

\paragraph{Experiment}
 
We split this dataset into 7 training shapes and 4 shapes for evaluation. Shapes in our test set are made of ``chickpea'' material, which exhibits the largest magnitude of granular surface deformation. We compare our method with 4 baselines namely, FARM~\cite{FARM}, BCICP~\cite{Ren2018}+ZoomOut~\cite{Melzi2019}, 3D-CODED~\cite{groueix2018b} and TransMatch~\cite{TransMatch}. Since the two data-driven baslines are not designed for training with key-point supervision, we use Equation~\ref{eqn:propogation} to generate dense-ground truth for training. In fairness, we report two variants of our method trained - one trained with key-point and the other with dense supervision denoted as ``Ours'' and ``Ours++'' respectively. We stress that this additional supervision is used only at training time while we maintain the test set to be fully-blind. We summarize our quantitative and qualitative results in Figure~\ref{fig:registration}. It is remarkable that our approach outperforms axiomatic and competing data-driven baselines by at least a two-fold margin. Importantly, despite our network sharing the same encoder~\cite{qi2016pointnet} as 3D-CODED~\cite{groueix2018b}, there is a striking difference in performance. We attribute this to our well-regularized deformation space.  

\subsection{Unsupervised Segmentation Transfer}
In this section, we demonstrate the generalization ability of our approach to model deformation between shapes with considerable topological differences. To that end, we consider the task of part-level segmentation over point clouds consisting of table and plane categories from ShapeNet~\cite{Yi2017LargeScale3S} dataset. Apart from topological differences and large structural variance, the absence of ground truth annotations exacerbates the challenge. In this setting, we compare our method with two Deep Implicit networks, namely DIF-Net~\cite{deng2021deformed} and DIT~\cite{zheng2021deep}, which model a volumetric deformation field between a learned template and training shapes. Our choice of baseline endows us with a fair ground of comparison between the two representations of the deformation field - MLP-based and ours.

\begin{figure*}[!t]
\begin{tabular}{cc}
\begin{minipage}{0.26\linewidth}
    \scriptsize
\begin{tabular}{llllll}
 & \multicolumn{2}{c}{Plane} &  \multicolumn{2}{c}{Table} \\ 
\cmidrule{2-3}\cmidrule{4-5}
 & \begin{tabular}[c]{@{}l@{}}CD\\ (x1e4)\end{tabular} & \begin{tabular}[c]{@{}l@{}}IoU\\ (\%)\end{tabular} & \begin{tabular}[c]{@{}l@{}}CD\\ (x1e4)\end{tabular} & \begin{tabular}[c]{@{}l@{}}IoU\\ (\%)\end{tabular} &  \\
 \cmidrule{1-5}
DIT~\cite{zheng2021deep} & 24.6 & 69.1 & 26.7 & 68.9 &  \\ 
\cmidrule{1-5}
DIF~\cite{deng2021deformed} & 15.0 & 78.0 & 11.2 & 79.3 &  \\ 
\cmidrule{1-5}
\begin{tabular}[c]{@{}l@{}}Ours \\(w/o Con)\end{tabular} & \textbf{0.5} & 88.8 & \textbf{2.6} & 88.9 &  \\ 
\cmidrule{1-5}
Ours & 0.6 & \textbf{89.3} & 3.0 & \textbf{90.0} & 
\end{tabular}%
\label{tab:unsup}
	\vspace{-0.2cm}
	
\end{minipage}
&
\hspace{1.5cm}

\begin{minipage}{0.6\linewidth}
	\scriptsize
	\vspace{0.2cm}
	\begin{overpic}[trim=0cm 0cm 0cm 0cm,clip, width=\linewidth]{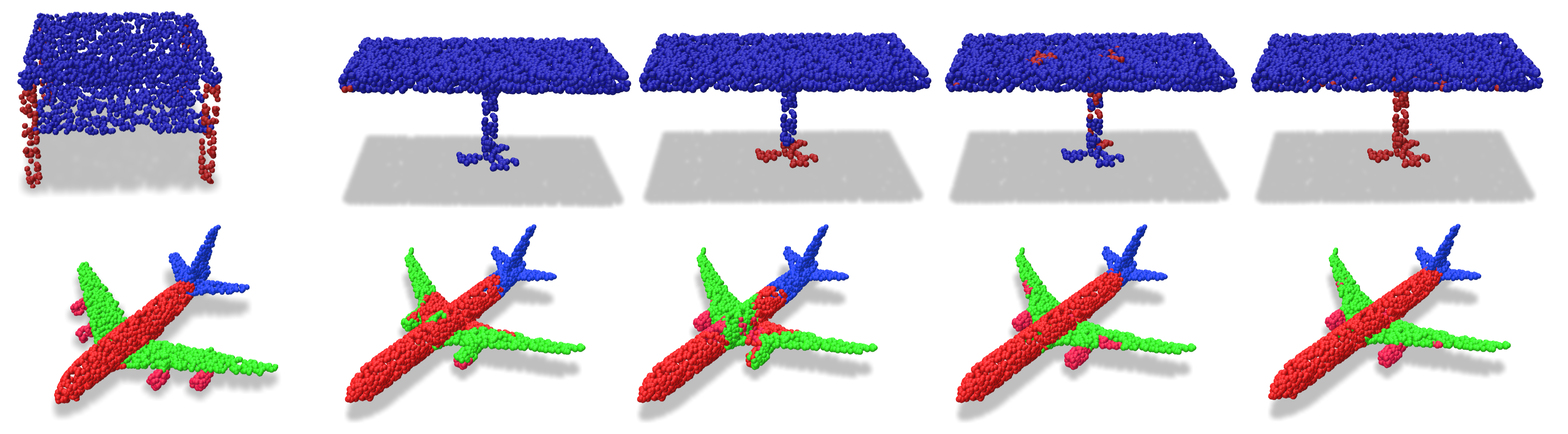}
		\put(5,-2){Source}
		\put(26,-2){DIT}
		\put(44,-2){DIF}
		\put(59,-2){Ours (w/o Con)}
		\put(83,-2){Ours}
	\end{overpic}
	\vspace{-0.3cm}
\end{minipage}

\end{tabular}

	\caption{ Ours (w/o Con) is a variant of our approach without any deformation constraints. Both of our variants show a significant improvement over baseline that models volumetric deformation fields. Qualitative result demonstrates color coded segmentation transfer across significant shape variability. }
	\label{fig:unsup}
\end{figure*}

\paragraph{Experiment and Discussion}
We train our approach using the unsupervised loss mentioned in Section~\ref{subsec:training} over 1000 objects sampled at random from each category. We consider 190 evaluation pairs per-category from the prescribed validation set and measure the segmentation accuracy by the IoU metric~\cite{Yi2017LargeScale3S}. In addition, we also measure the bi-directional Chamfer's distance of reconstructed geometries. We summarize our quantitative observation along with a qualitative example in Figure~\ref{fig:unsup}. We remark that while deformation fields between aforementioned categories are not strictly volume-preserving, we still observe a noticeable improvement over the baseline. This is because our deformation priors help in \emph{structuring} the space of deformations, which explicitly avoids degeneracy such as collapsing shape parts. This remark is corroborated by a lower (preferred) Chamfer distance while there is \emph{a decline} in the accuracy when no regularization is applied.

\begin{figure}[!h]
\centering
     \begin{subfigure}[b]{0.9\textwidth}
         \centering
    \includegraphics[width=\textwidth]{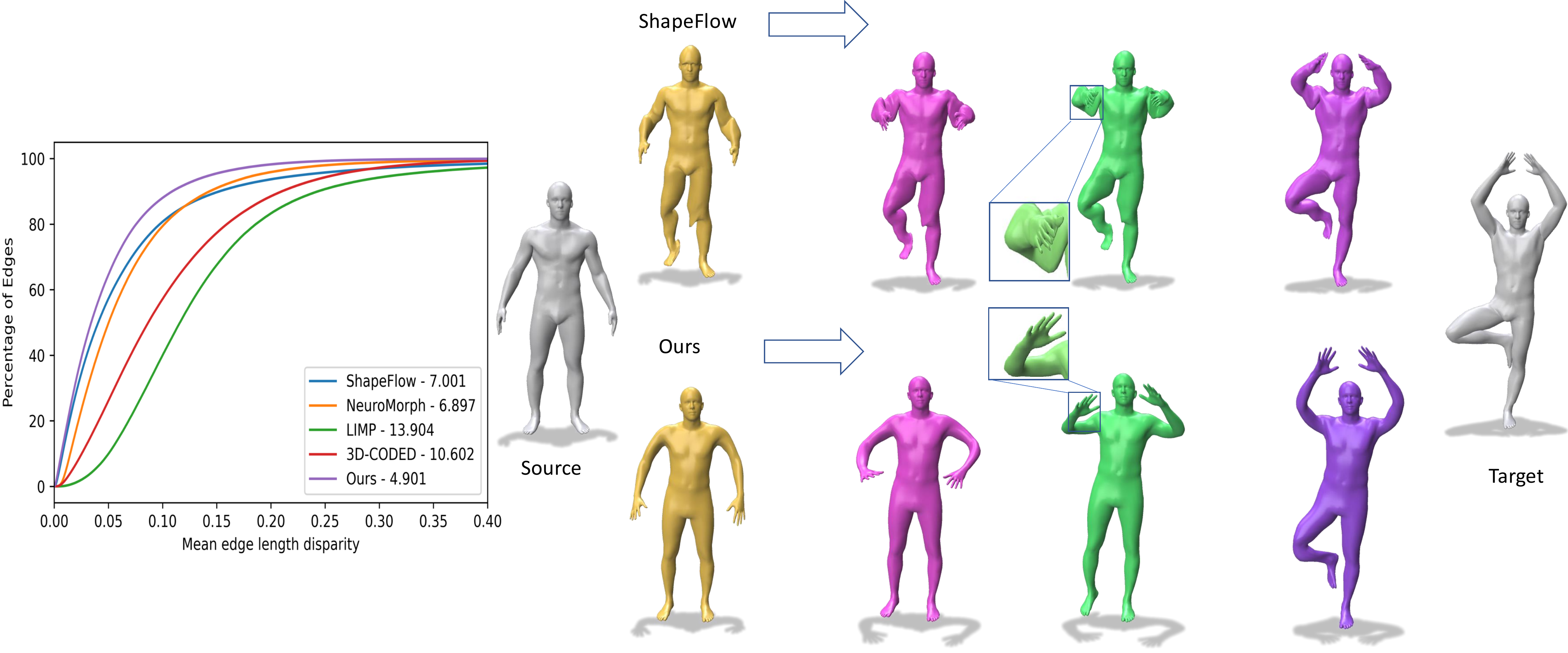}
     \end{subfigure}
     \caption{Quantitative and qualitative comparison of interpolation. While ShapeFlow~\cite{jiang2020shapeflow} enforces volume preservation prior, its latent deformation space are not distortion-free. By better structuring the latent-space (c.f Eqn~\ref{eqn:lossZ}), our sampled intermediate shapes are near distortion-free. }
     \label{fig:interp}
\end{figure}

\subsection{Shape Interpolation}

A notable characteristic of a well-structured latent space is the ability to produce plausible intermediate shapes given a source and a target. This task is commonly referred to as shape interpolation. Since there exists no canonical path, interpolation sequences are gauged by the extent to which intrinsic metrics are preserved, in particular isometric distortion~\cite{kmp_shape_space_sig_07}. For this setting, we consider the FAUST~\cite{bogo2014faust} dataset, where, we train our method on the first 80 shapes and evaluate over the last 20 shape pairs. We compare our method against four baselines namely 3D-CODED~\cite{groueix20183d}, NeuroMorph~\cite{neuromorph}, LIMP~\cite{Cosmo2020} and ShapeFlow~\cite{jiang2020shapeflow}. Since LIMP employs a fixed-size decoder and NeuroMorph uses a separate interpolation module involving an explicit computation of correspondence matrix, both of these approaches are limited by shape resolution. On the other hand, our approach is resolution agnostic and outperforms the baselines by a discernible margin as summarized in Figure~\ref{fig:interp}. This improvement over the baseline is due to the incorporation of our latent deformation priors in a computationally feasible manner, which we will be justified through an extensive ablation study in the supplementary.

\subsection*{Additional Results}
 In addition to the results shown above, we also present qualitative correspondence results between neural implicit fields and real-world data in the supplementary. More specifically, in Section 5 of the supplementary, we show qualitative interpolation and correspondence results between implicitly defined surfaces.  Then, in Section 6.1 of supplementary, we show qualitative correspondence results in the form of texture transfer between pair of shapes from the CMU-Panoptic dataset~\cite{Joo_2017_TPAMI} consisting of point clouds acquired from from \emph{Kinect RGB-D sensor}. Finally, in Section 6.2, we also show the versatility of our representation in modelling deformation field between shapes that have more freedom regarding such as meshes of the human heart~\cite{Strocchi2020}.

\section{Conclusion, Limitations and Future Work}
\label{sec:conclusion}

We presented an effective representation of deformation fields, which allows learning a \textit{reduced set} of shape-specific deformation parameters while constructing the continuous deformation field using mesh-free approximation. A key observation behind our method is that in many settings, the space of realistic deformations is well-constrained and expressed with a small set of parameters. To that end, we demonstrated that our approach can achieve significant improvement upon existing baselines across challenging downstream applications and remarkably reduce the dependence on training data. Moreover, this representation also endowed us with access to first-order derivatives in closed form, thereby facilitating the use of strong first-order regularization. Our approach still has some limitations and leads to possible exciting future work. Firstly, while our approach produces a smooth deformation field in principle, there is no guarantee of bijectivity or invertability. Second, instead of fixed nodal positions, optimizing with respect to our approximation function would also be an interesting direction to study. 

\textbf{Acknowledgments: } Experiments were performed using HPC resources from GENCI-IDRIS (Grant 2021-AD011013104). Parts of this work were supported by the ERC Starting Grants No. 758800 (EXPROTEA) and No. 802554 (SPECGEO), the ANR AI Chair AIGRETTE, an Alexander von Humboldt Foundation Research Fellowship. We thank Robin Magnet and Gautam Pai for their feedback in improving our manuscript.
\clearpage
\section{Supplementary}
\label{sec:supplementary}
\pdfoutput=1
In Section~\ref{sec:implDetails} we provide the implementation details of our novel deformation field representation, followed by a quantitative comparison with potential alternative representations in Section~\ref{sec:cmpAlt}. Then, we perform an extensive ablation study in Section~\ref{sec:ablStudy} to justify the need for regularization and our design choices. In Section~\ref{sec:supEffect}, we quantitatively show the reduced need for supervision of our approach by comparing with relevant baselines. In Section~\ref{sec:unusupImplicit}, we further highlight the generalization ability of our reduced representation in establishing high-quality correspondences between learnt implicitly defined surfaces. Finally in Section ~\ref{sec:realWorlddata}, we demonstrate the robustness of our approach in estimating correspondence between real-world data acquired from RGB-D sensor and scans of human hearts.

\subsection{Implementation Details}
\label{sec:implDetails}
\subsubsection{Pre-Processing}
We scale all training shapes to fit into a unit sphere including our template. Then, we sample nodes from within the volume defined by the template if our shape collection is a mesh. In case of point cloud, we simply augment by adding random Gaussian noise. Once sampled, we fix the positions of nodes. Finally, we pre-compute $\Phi$ and $\nabla_{x,y,x}\Phi$ at each evaluation points. Please note this pre-computation is performed only to accelerate training when evaluation points are known and it is not a strict requirement. 

\subsubsection{Closed-form expression for deformation field gradient}
We show the pre-computation of $\nabla_{x,y,x}\Phi$ in this subsection. From Equation 4 of the main paper, the Jacobian of the deformation field is given as,

\begin{equation*}
\mathbb{J} = \frac{\partial \mathbf{u}(\mathbf{x})}{\partial \mathbf{x}_{(d)}} = \sum_{i=1}^{K} \frac{\partial \Phi_{i}(\mathbf{x})}{\partial \mathbf{x}_{(d)}} u_{i} \hspace{4mm} \text{  where, }\ \mathbf{x}_{(d)} = [x, y, z]^{T}
\end{equation*}

Expanding $\frac{\partial \Phi_{i}(\mathbf{x})}{\partial \mathbf{x}_{(d)}}$,

\begin{equation*}
\begin{aligned}
 \frac{\partial \Phi_{i}(\mathbf{x})}{\partial \mathbf{x}_{(d)}} &= \frac{\partial \left(p^{T}(\mathbf{x})[M(\mathbf{x})]^{-1} w_{i}(\mathbf{x}) p\left(\mathbf{q}_{i}\right) \right)}{\partial \mathbf{x}_{(d)}} \\
&= \frac{\partial p^{T}(\mathbf{x})}{\partial \mathbf{x}_{(d)}} [M(\mathbf{x})]^{-1} w_{i}(\mathbf{x}) p\left(\mathbf{q}_{i}\right) + p^{T}(\mathbf{x}) \frac{\partial [M(\mathbf{x})]^{-1}} {\partial \mathbf{x}_{(d)}} w_{i}(\mathbf{x}) p\left(\mathbf{q}_{i}\right) + p^{T}(\mathbf{x}) [M(\mathbf{x})]^{-1} \frac{\partial w_{i}(\mathbf{x})}{\partial \mathbf{x}_{(d)}} p\left(\mathbf{q}_{i}\right)
\end{aligned}
\end{equation*}

Using a $1^{st}$ order polynomial basis $p(\mathbf{x})=\left[1\ x\ y\ z\right]^T$ and the fact that
$\frac{\partial [M(\mathbf{x})]^{-1}} {\partial \mathbf{x}}=-[M(\mathbf{x})]^{-1}\left(\partial M / \partial \mathbf{x}_{(k)}\right) [M(\mathbf{x})]^{-1}$

\begin{equation*}
\begin{aligned}
\frac{\partial \Phi_{i}(\mathbf{x})}{\partial \mathbf{x}_{(d)}} &= [M(\mathbf{x})]^{-1} w_{i}(\mathbf{x}) p\left(\mathbf{q}_{i}\right) - p^{T}(\mathbf{x}) [M(\mathbf{x})]^{-1} \frac{\partial M(\mathbf{x})} {\partial \mathbf{x}_{(d)}} [M(\mathbf{x})]^{-1} w_{i}(\mathbf{x}) p\left(\mathbf{q}_{i}\right) + p^{T}(\mathbf{x}) [M(\mathbf{x})]^{-1} \frac{\partial w_{i}(\mathbf{x})}{\partial \mathbf{x}_{(d)}} p\left(\mathbf{q}_{i}\right)
\end{aligned}
\end{equation*}

From the definition of $M(\mathbf{x})$

\begin{equation*}
    \frac{\partial M(\mathbf{x})} {\partial \mathbf{x}_{(d)}} = \sum_{i=1}^{K} p\left(\mathbf{q}_{i}\right) p^{T}\left(\mathbf{q}_{i}\right)  \frac{\partial w_{i}(\mathbf{x})} {\partial \mathbf{x}_{(d)}}
\end{equation*}

Where,  
\begin{equation*}
    \frac{\partial w_{i}(\mathbf{x})} {\partial \mathbf{x}_{(d)}} = 
    \begin{dcases}
    \frac{\left(-3 \cdot\left(1-||\mathbf{x}-\mathbf{q}_i||_2 / r_i\right)^{2}\right)} {\left(r_i \cdot ||\mathbf{x}-\mathbf{q}_i||_2\right)} \cdot \left(\mathbf{x}-\mathbf{q}_i\right)& \text{if } ||\mathbf{x} - \mathbf{q}_i||^{2}_{2}\leq r_i\\
    0,              & \text{otherwise}
    \end{dcases}
\end{equation*}

Therefore, $\forall \mathbf{q}_i \ \ s.t. ||\mathbf{x} - \mathbf{q}_i||^{2}_{2}\leq r_i$
\begin{equation}
\begin{aligned}
\frac{\partial \Phi_{i}(\mathbf{x})}{\partial \mathbf{x}_{(d)}} &= [M(\mathbf{x})]^{-1} w_{i}(\mathbf{x}) p\left(\mathbf{q}_{i}\right)\\
-&p^{T}(\mathbf{x}) [M(\mathbf{x})]^{-1} \sum_{i=1}^{K} p\left(\mathbf{q}_{i}\right) p^{T}\left(\mathbf{q}_{i}\right) \frac{\left(-3 \cdot\left(1-||\mathbf{x}-\mathbf{q}_i||_2 / r_i\right)^{2}\right)} {\left(r_i \cdot ||\mathbf{x}-\mathbf{q}_i||_2\right)} \cdot \left(\mathbf{x}-\mathbf{q}_i\right)(\mathbf{x}) p\left(\mathbf{q}_{i}\right) \\
&+ p^{T}(\mathbf{x}) [M(\mathbf{x})]^{-1} \frac{\left(-3 \cdot\left(1-||\mathbf{x}-\mathbf{q}_i||_2 / r_i\right)^{2}\right)} {\left(r_i \cdot ||\mathbf{x}-\mathbf{q}_i||_2\right)}  \left(\mathbf{x}-\mathbf{q}_i\right) p\left(\mathbf{q}_{i}\right)
\end{aligned}
\label{eqn:fullExpansion}
\end{equation}

From Equation\ref{eqn:fullExpansion}, we see that $\mathbb{J}$ is independent of the deformation parameter $u_i$. Since we know the node positions $\mathbf{q}_i$, in scenarios where evaluation points $\mathbf{x}$ are known, $\mathbb{J}$ can be pre-computed.

\subsubsection{Architecture}
\begin{figure}[h]
\centering
\includegraphics[width=\textwidth]{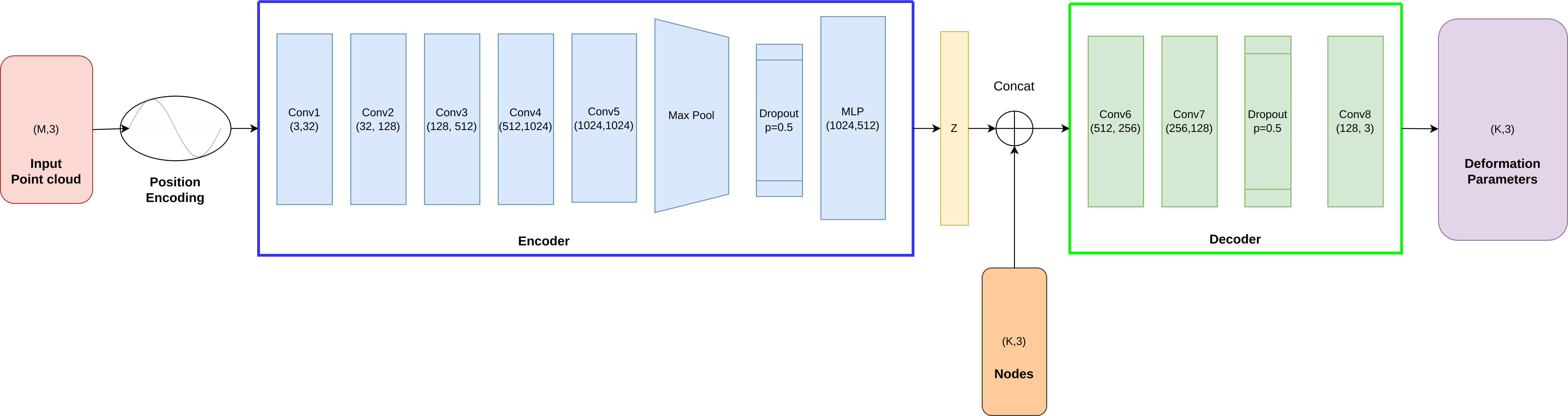}
\caption{A detailed summary of our architecture}
\label{fig:archDetailed}
\end{figure}

A detailed overview of our architecture is show in Figure~\ref{fig:archDetailed}. We apply position encoding to our input coordinates following Tanick~\etal~\cite{tancik2020fourfeat}. Input coordinates are embedded to the surface of a 128 dimensional hypersphere.

\subsubsection{Hyper-Parameters}
In this section, we provide details on hyper-parameters, choice of template and nodes corresponding to each experiment in our main paper. 

\paragraph{Shape Correspondence and Registration}

\begin{figure}[h]
\centering
\includegraphics[width=0.8\textwidth]{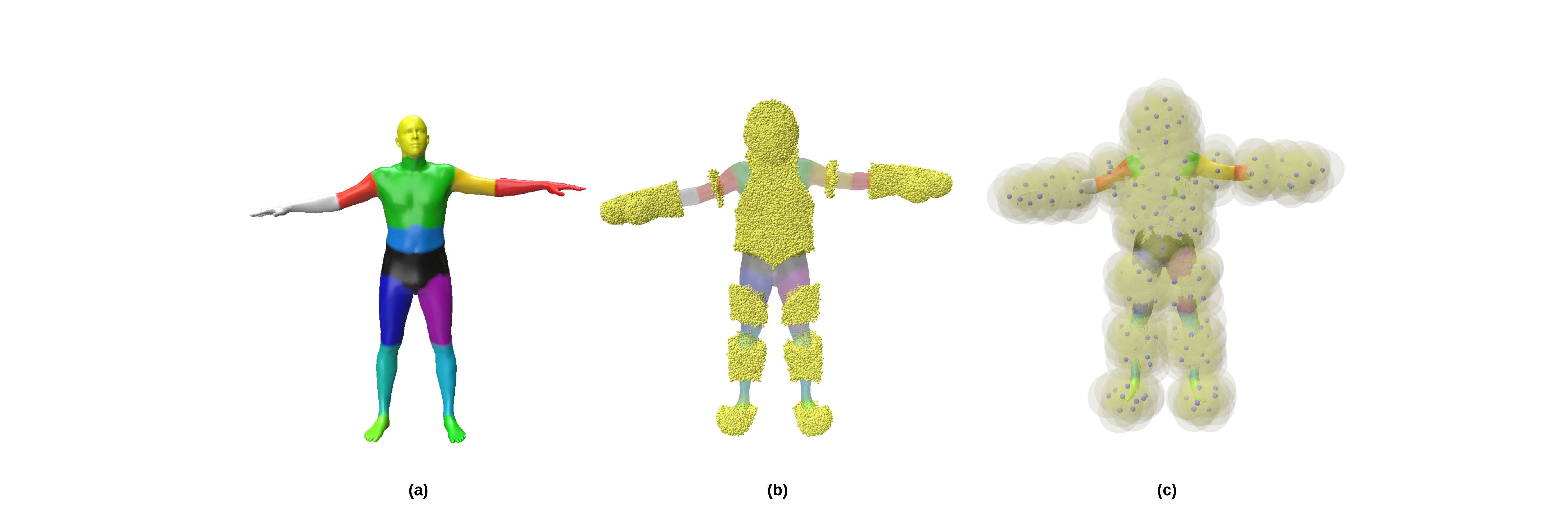}
\caption{Different stages of node sampling strategy. (a) First, we segment the template using~\cite{SMPL2015}. (b) Second, we sample points close to the surface of the template mesh called candidates. (c) Finally, we sub-sample from candidates until all vertices have 4 non-planar nodes in its vicinity. Nodes are shown as blue points with its region of influence in transparent yellow. }
\label{fig:nodeSampling}
\end{figure}

We discuss the hyper-parameters used in Experiments 5.1 and 5.2 of the main paper respectively. Since our template is a mesh in these two experiments, we leverage the connectivity while sampling nodes in order to eliminate a node from influencing two vertices that belong to different semantic regions as shown in Figure~\ref{fig:nodeSampling}. Hyper-parameters used during training and inference are given in Table~\ref{tab:exp1HP}

\begin{table}[H]
\centering
\resizebox{0.7\textwidth}{!}{%
\begin{tabular}{@{}|c|cccccc|ccc|@{}}
\toprule
Mode &
  \multicolumn{6}{c|}{Training} &
  \multicolumn{3}{c|}{Inference} \\ \midrule
Variable &
  \multicolumn{1}{c|}{K} &
  \multicolumn{1}{c|}{$\lambda_1$} &
  \multicolumn{1}{c|}{$\lambda_2$} &
  \multicolumn{1}{c|}{$\lambda_3$} &
  \multicolumn{1}{c|}{$\lambda_4$} &
  $r_i$ &
  \multicolumn{1}{c|}{$\Lambda_1$} &
  \multicolumn{1}{c|}{$\Lambda_2$} &
  $\Lambda_3$ \\ \midrule
Value &
  \multicolumn{1}{c|}{300} &
  \multicolumn{1}{c|}{1} &
  \multicolumn{1}{c|}{5e-3} &
  \multicolumn{1}{c|}{1e-2} &
  \multicolumn{1}{c|}{5e-3} &
  2e-1 &
  \multicolumn{1}{c|}{1} &
  \multicolumn{1}{c|}{1e-4} &
  1e-3 \\ \bottomrule
\end{tabular}%
}
\caption{Hyper-parameters used for our non-rigid shape correspondence and registration experiments}
\label{tab:exp1HP}
\end{table}

\begin{figure}[h]
\centering
\includegraphics[width=0.8\textwidth]{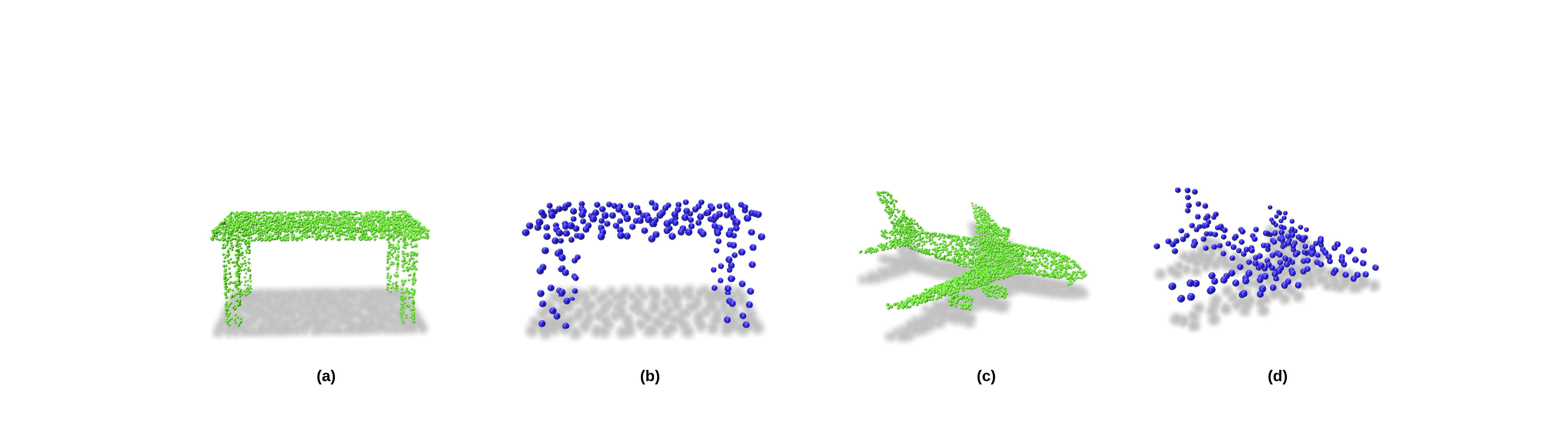}
\caption{Template point clouds corresponding to Table and Plane class used in Experiment 5.3 from the main paper. (a) and (c) denotes the template while (b) and (d) denote sampled nodes.}
\label{fig:unsupTemplates}
\end{figure}

\paragraph{Unsupervised Segmentation Transfer}
The template point cloud and corresponding sampled nodes are shown in Figure~\ref{fig:unsupTemplates}. Since the shape collection exhibit significant structural difference, we slightly relax the first-order regularization. We perform test-time refinement for 200 steps. Our hyper-parameters are summarized in Table~\ref{tab:exp2Hparam}.

\begin{table}[H]
\centering
\resizebox{0.7\textwidth}{!}{%
\begin{tabular}{@{}|c|cccccc|ccc|@{}}
\toprule
Mode &
  \multicolumn{6}{c|}{Training} &
  \multicolumn{3}{c|}{Inference} \\ \midrule
Variable &
  \multicolumn{1}{c|}{K} &
  \multicolumn{1}{c|}{$\lambda_1$} &
  \multicolumn{1}{c|}{$\lambda_2$} &
  \multicolumn{1}{c|}{$\lambda_3$} &
  \multicolumn{1}{c|}{$\lambda_4$} &
  $r_i$ &
  \multicolumn{1}{c|}{$\Lambda_1$} &
  \multicolumn{1}{c|}{$\Lambda_2$} &
  $\Lambda_3$ \\ \midrule
Value &
  \multicolumn{1}{c|}{300} &
  \multicolumn{1}{c|}{1} &
  \multicolumn{1}{c|}{1e-4} &
  \multicolumn{1}{c|}{1e-3} &
  \multicolumn{1}{c|}{1e-4} &
  2e-1 &
  \multicolumn{1}{c|}{1} &
  \multicolumn{1}{c|}{1e-4} &
  1e-3 \\ \bottomrule
\end{tabular}%
}
\caption{Hyper-parameters used in unsupervised segmentation transfer experiment.}
\label{tab:exp2Hparam}
\end{table}

\paragraph{Shape Interpolation}
We do not perform test-time refinement for this experiment. Table~\ref{tab:exp3Hparam} summarizes the training hyper-parameters. We use the same nodes as in shape correspondence experiment.

\begin{table}[H]
\centering
\resizebox{0.5\textwidth}{!}{%
\begin{tabular}{@{}|c|cccccc|@{}}
\toprule
Mode &
  \multicolumn{6}{c|}{Training} \\ \midrule
Variable &
  \multicolumn{1}{c|}{K} &
  \multicolumn{1}{c|}{$\lambda_1$} &
  \multicolumn{1}{c|}{$\lambda_2$} &
  \multicolumn{1}{c|}{$\lambda_3$} &
  \multicolumn{1}{c|}{$\lambda_4$} &
  $r_i$ \\ \midrule
Value &
  \multicolumn{1}{c|}{300} &
  \multicolumn{1}{c|}{1} &
  \multicolumn{1}{c|}{1e-4} &
  \multicolumn{1}{c|}{1e-2} &
  \multicolumn{1}{c|}{1} &
  2e-1 \\ \bottomrule
\end{tabular}%
}
\caption{Hyper-parameters used in shape interpolation experiment.}
\label{tab:exp3Hparam}
\end{table}

\subsection{Comparison with Alternatives}
\label{sec:cmpAlt}
We perform quantitative comparison between our approach and plausible alternatives in terms of accuracy and computational efficiency. We compare with two main alternative representations. First, deformation at each point in the template shape is predicted using a point-wise MLP, abbreviated as PW-MLP. Second, we compare with an interpolation variant, where we replace the mesh-free approximation with Radial Basis Function (RBF) interpolation. We provide a brief overview of this interpolation variant below. In order to have a controlled setting, all our experiments are performed on a machine with Nvidia Ampere A100 GPUs and AMD 7302 3Ghz CPU over same training, batch-size and number of points used as input to the encoder.

\paragraph{RBF Interpolation}

Instead of \emph{approximating} the deformation field across the surface using mesh-free functions, we use RBF interpolation to \emph{interpolate} the deformation field based on observations at nodes. Our main pipeline (as shown in Figure.1 of main paper) remains the same, barring the fact that, here, we interpolate the deformation field. Let $U$ be the predicted deformation parameters at nodes $\mathcal{Q}$, $\varphi(\cdot)$ be the RBF kernel, $\Phi$ to be its matrix representation, interpolant function $f(\cdot)$ at a point $\mathbf{x} \in \mathbb{R}^{3}$ is given as:
\begin{equation}
f(x) = \sum_{i=1}^{K} \Phi^{-1} U \varphi\left(\mathbf{x}, \mathbf{q}_i\right).
\label{eqn:rbfInterp}
\end{equation}
Here $\mathbf{q}_i \in \mathcal{Q}$ and K denotes the total number of nodes. The kernel function $\varphi(\cdot)$~\cite{Hardy1971} and its matrix representation $\Phi$ are defined as:
\begin{equation}
\Phi_{mn} \coloneqq \varphi(\mathbf{q}_m, \mathbf{q}_n)=\sqrt{C + \epsilon_0 ||\mathbf{q}_m - \mathbf{q}_n||^{2}},
\label{eqn:interpolant}
\end{equation}
where, $m$ and $n$ denote the $m^{th}$ row and $n^{th}$ column of $\Phi$ respectively, $\mathbf{q}_n, \mathbf{q}_n \in \mathcal{Q}$. Please note that $\Phi$ is positive definite by its construction. $C=1$ and $\epsilon_0=50$ are taken to be constants. Similar to the case of mesh-free approximation, it is easy to see that evaluation of Jacobian is independent of values of deformation field itself:
\begin{equation}
    \mathbb{J} = \nabla_{x,y,z}f(\mathbf{x}) = \Phi^{-1} U \left(\sum_{i=1}^{K} \left[\frac{\partial \varphi\left(\mathbf{x}, \mathbf{q}_i\right)}{\partial x}, \frac{\partial \varphi\left(\mathbf{x}, \mathbf{q}_i\right)}{\partial y}, \frac{\partial \varphi\left(\mathbf{x}, \mathbf{q}_i\right)}{\partial z}\right]\right)
\label{eqn:gradRBF}
\end{equation}

\paragraph{Interpolation vs Approximation}
Although interpolation is close to our proposed representation in terms of effective reduction, there are three key differences between both representations. First, the weighting function expressed through $\varphi(\cdot)$ has an \emph{infinite support} while our approximation has compact support. Second, our representation guarantees the approximation function of $n^{th}$ order consistency depending on the polynomial basis. Third, in the case of interpolation, $f(\mathbf{q}_i) = u_i | u_i \in U$, whereas while approximating, $u(\mathbf{q}_i) \neq u_i | u_i \in U$. This distinction is important from the meaning it endows our network $\mathcal{F}(\cdot)$. In the interpolation case, it amounts to predicting the \emph{deformation field} whereas in the approximation case, $\mathcal{F}(\cdot)$ predicts \emph{deformation parameters}.

\begin{figure}[h]
\centering
\includegraphics[width=0.9\textwidth]{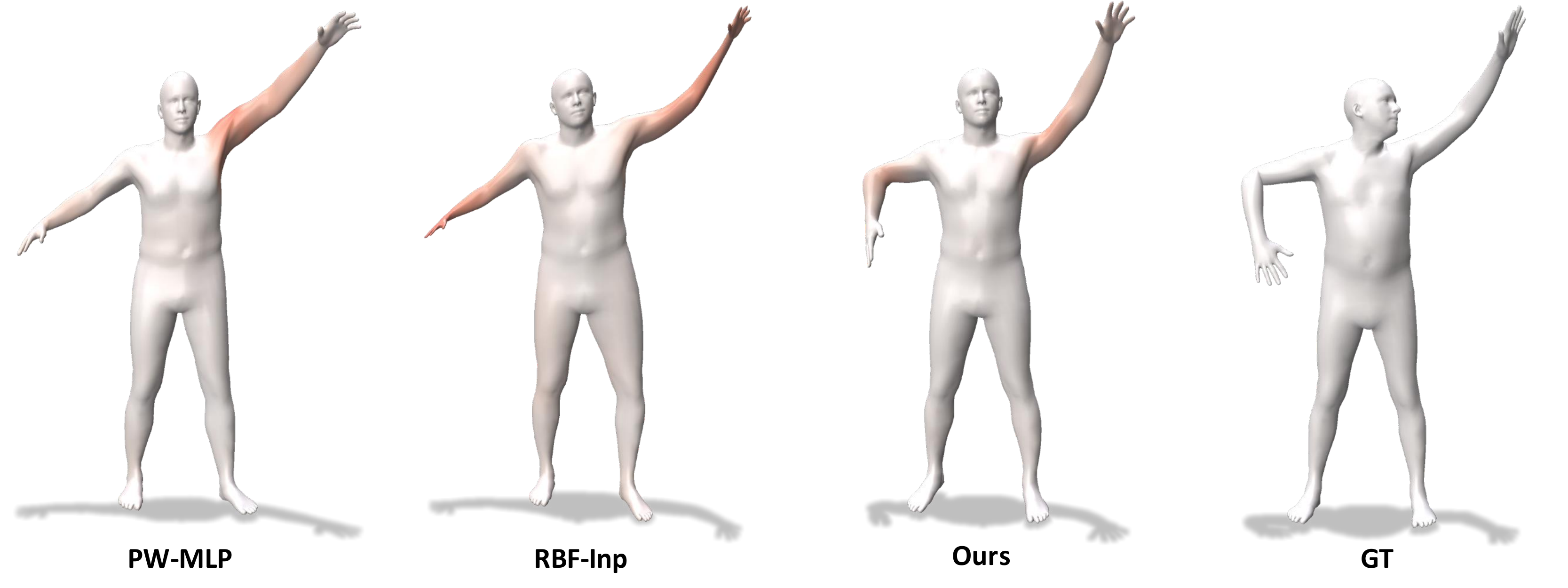}
\caption{Qualitative comparison between Point-wise MLP(PW-MLP), RBF interpolation (RBF-Inp) and our approach (Ours). Meshes are color-coded with area distortion. Increasing shades of red signifies larger distortion.}
\label{fig:figNRSC}
\end{figure}

\paragraph{Discussion}
We summarize the quantitative and qualitative results in Table~\ref{tab:NRSC}. Our approach shows nearly $325\times$ improvement in speed when applying first-order regularization compared to the PW-MLP. While both reduced variants, namely, interpolation and approximation show comparable timings, there is, however, a significant performance difference between them in terms of correspondence accuracy. We attribute this observation mainly to the compactness of our approximation. Without such compactness, deformation fields corresponding to different semantic regions may influence one another. For instance, the deformation field corresponding to left arm of a human can affect the way the right arm moves or the ``pull-effect''. This particular example is characterized in Figure~\ref{fig:figNRSC}, where \emph{interpolating} the deformation field fails to capture the articulation. In addition, the ``pull-effect'' also shortens the length of the left-arm in comparison to the right.

\begin{table}[]
\centering
\resizebox{\textwidth}{!}{%
\begin{tabular}{@{}|l|lll|lll|lll|@{}}
\toprule
Constraint &
  \multicolumn{3}{l|}{None} &
  \multicolumn{3}{l|}{$\mathcal{L}_{vol} + \mathcal{L}_{arap}$} &
  \multicolumn{3}{l|}{$\mathcal{L}_{vol} + \mathcal{L}_{arap} + \mathcal{L}_{Z}$} \\ \midrule
Method &
  \multicolumn{1}{l|}{PW-MLP} &
  \multicolumn{1}{l|}{RBF-Inp} &
  Ours &
  \multicolumn{1}{l|}{PW-MLP} &
  \multicolumn{1}{l|}{RBF-Inp} &
  Ours &
  \multicolumn{1}{l|}{PW-MLP} &
  \multicolumn{1}{l|}{RBF-Inp} &
  Ours \\ \midrule
\multicolumn{1}{|c|}{Time(iter/ms)} &
  \multicolumn{1}{l|}{32.0} &
  \multicolumn{1}{l|}{8.0} &
  7.8 &
  \multicolumn{1}{l|}{250.4} &
  \multicolumn{1}{l|}{8.2} &
  8.0 &
  \multicolumn{1}{l|}{4098.3} &
  \multicolumn{1}{l|}{12.6} &
  \textbf{12.1} \\ \midrule
SCAPE (PC+N) &
  \multicolumn{1}{l|}{13.7} &
  \multicolumn{1}{l|}{9.9} &
  9.8 &
  \multicolumn{1}{l|}{12.6} &
  \multicolumn{1}{l|}{9.1} &
  7.0 &
  \multicolumn{1}{l|}{14.8} &
  \multicolumn{1}{l|}{9.3} &
  \textbf{6.6} \\ \midrule
SHREC19 &
  \multicolumn{1}{l|}{7.4} &
  \multicolumn{1}{l|}{8.5} &
  7.7 &
  \multicolumn{1}{l|}{7.1} &
  \multicolumn{1}{l|}{8.0} &
  5.2 &
  \multicolumn{1}{l|}{9.1} &
  \multicolumn{1}{l|}{8.1} &
  \textbf{4.8} \\ \bottomrule
\end{tabular}%
}
\caption{Quantitative comparison of efficiency and correspondence accuracy between possible alternative representations and our approach. Imposing latent-space regularization results in prohibitive computation effort using a standard PW-MLP representation of deformation fields. }
\label{tab:NRSC}
\end{table}

\subsection{Ablation Study}
\label{sec:ablStudy}


In this section, we perform an in-depth ablation study to analyze the effect of first-order regularization during training and inference. Then, we ablate our choice of number of nodes, its influence, different positioning strategy and pose of template used. Training regularization (Tr-Regularization) and Test-time regularization (Te-Regularization) refers to Equation 6 and Equation 11 from the main paper respectively. We re-train all methods on the same 1000 SURREAL shapes~\cite{varol17_surreal} mentioned in the main paper while evaluating them on SHREC'19~\cite{Melzi2019} and SCAPE (PC+N) datasets for the non-rigid shape correspondence task. 

\begin{table}[H]
\centering
\resizebox{0.7\textwidth}{!}{%
\begin{tabular}{@{}|c|cccc|cc|c|@{}}
\toprule
\multirow{2}{*}{Dataset} & \multicolumn{4}{c|}{Tr-Regularization} & \multicolumn{2}{c|}{Te-Regularization} & Ours \\ \cmidrule(l){2-8} 
 & \multicolumn{1}{c|}{None} & \multicolumn{1}{c|}{$\mathcal{L}_{arap}$} & \multicolumn{1}{c|}{$\mathcal{L}_{vol}$} & $\mathcal{L}_{vol} + \mathcal{L}_{arap}$ & \multicolumn{1}{c|}{None} & $\mathcal{L}_{CD}$ & All \\ \midrule
SHREC’19 & \multicolumn{1}{c|}{7.7} & \multicolumn{1}{c|}{6.9} & \multicolumn{1}{c|}{7.3} & 5.2 & \multicolumn{1}{c|}{7.9} & 5.0 & \textbf{4.8} \\ \midrule
SCAPE & \multicolumn{1}{c|}{9.8} & \multicolumn{1}{c|}{10.2} & \multicolumn{1}{c|}{8.7} & 7.0 & \multicolumn{1}{c|}{12.5} & 6.7 & \textbf{6.6} \\ \bottomrule
\end{tabular}%
}
\caption{Ablation study on regularization at training and test time. All training regularization are imposed when ablating test-time regularization and vice-versa.  }
\label{tab:ablation}
\end{table}

\begin{figure}[h]
\centering
\includegraphics[width=\textwidth]{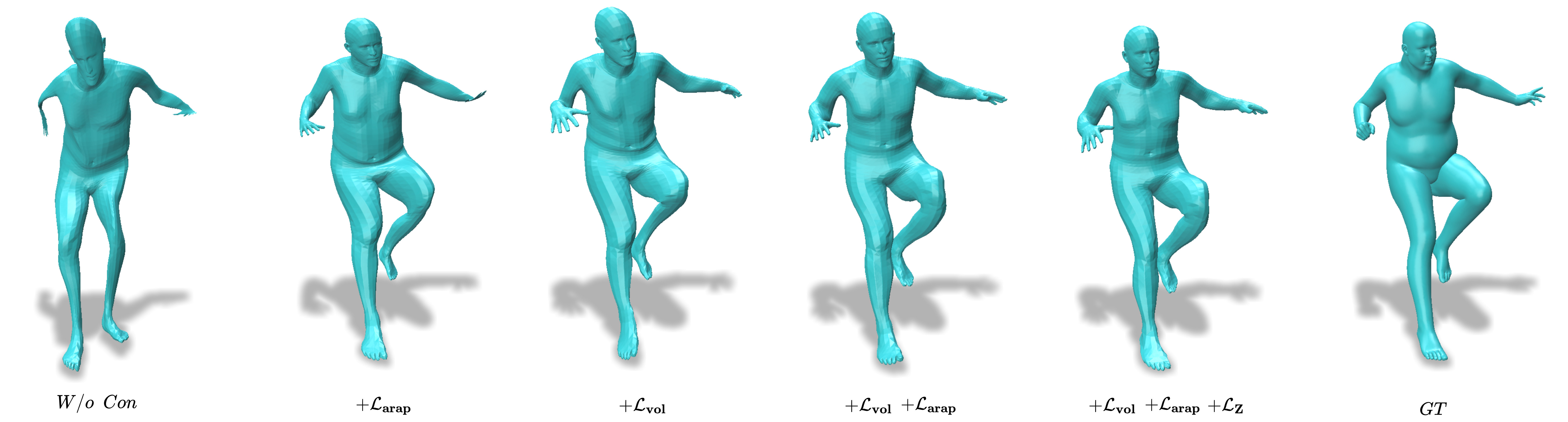}
\caption{Effect of different regularization applied to the deformation field. While enforcing $\mathcal{L}_{arap}$ reconstructs the shape, the lack of $\mathcal{L}_{vol}$ leads to ``collapse'' effect at hands and legs. Similarly using $\mathcal{L}_{vol}$ does not result in a distortion-free reconstruction. Finally, incorporating $\mathcal{L}_Z$ produces distortion-free deformation at joints (see right elbow).}
\label{fig:ablationQual}
\end{figure}

\subsubsection{Training Regularization}
Our main observations on the efficacy of deformation field regularization is summarized in Table~\ref{tab:ablation}. Although the mapping between the template shape and target are highly non-isometric, yet incorporating first-order regularization show a notable improvement in accuracy. 

This is because we do not restrict ourselves to exactly volume preserving deformations, but rather use our regularizers to penalize implausible deformations, that can  incur significant volume distortion. We empirically observe such regularization helps in producing better results especially in the presence of limited training data. We also show an example of deformed templates corresponding to the losses we ablate in Figure~\ref{fig:ablationQual}. While all variants of our method recover the pose of the template, the reconstructions are plausible only when enforcing first-order regularization.

\begin{figure}[H]
\centering
\includegraphics[width=\textwidth]{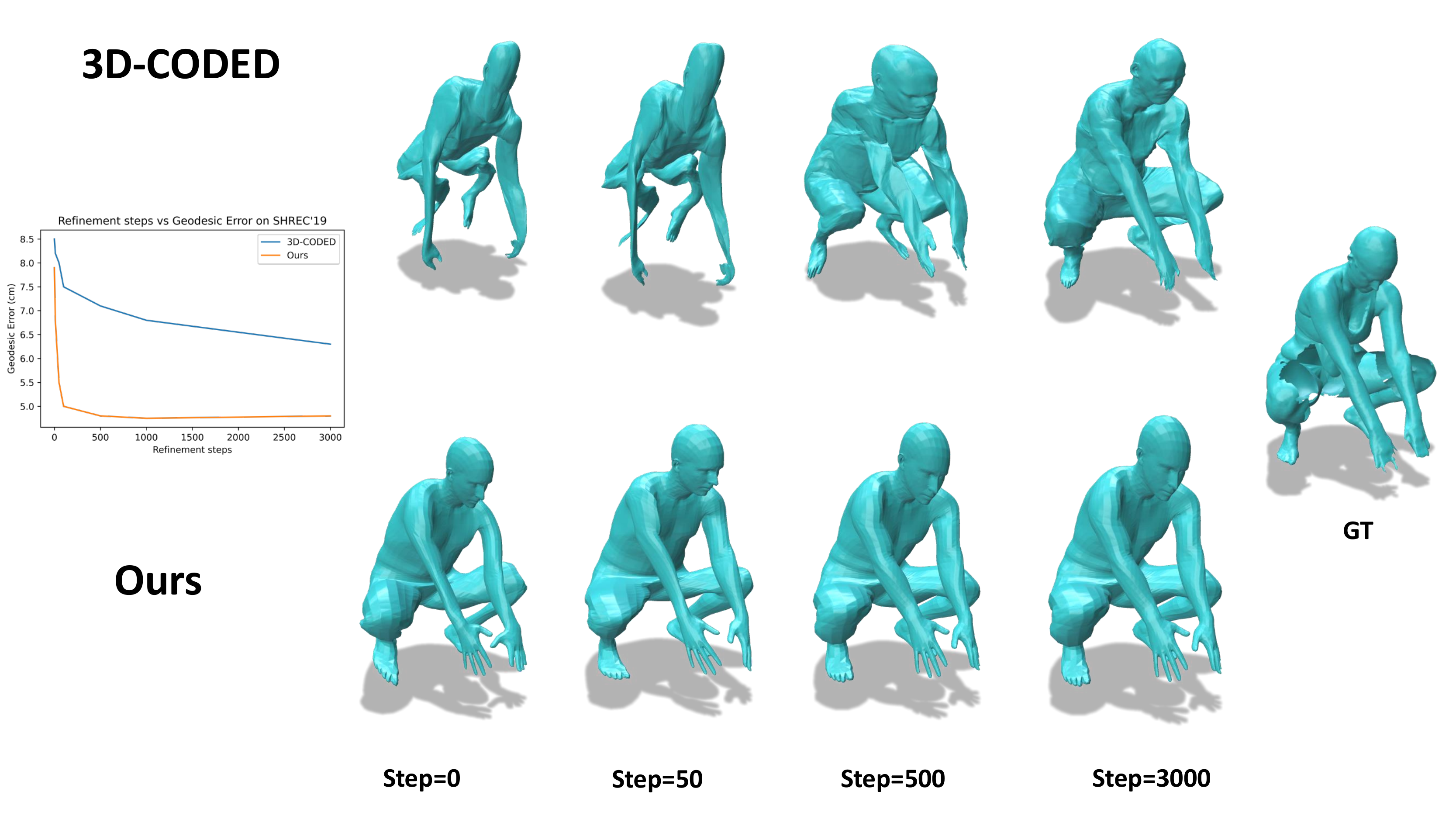}
\caption{Quantitative and qualitative illustration of test-time refinement. Our demonstrates a plausible reconstruction w/o refinement which baseline fails to accomplish. This results in requiring significantly less test-time refinement efforts as compared to 3D-CODED.}
\label{fig:TeRefinement}
\end{figure}

\subsubsection{Test-time Regularization}
We perform test-time refinement to enhance the reconstruction similar to 3D-CODED~\cite{groueix20183d}. However, owing to our structured deformation space, our method provides a more plausible initialization, thereby requiring significantly less refinement steps as shown in Figure~\ref{fig:TeRefinement}. This is particularly beneficial in expediting the inference process.

\begin{table}[H]
\centering
\resizebox{0.7\textwidth}{!}{%
\begin{tabular}{|c|c|cc|ccc|ccc|}
\hline
\# Nodes & 100 & \multicolumn{2}{c|}{300} & \multicolumn{3}{c|}{900} & \multicolumn{3}{c|}{2700} \\ \hline
Radius   & 0.5 & \multicolumn{1}{c|}{0.2} & 0.5 & \multicolumn{1}{c|}{0.1} & \multicolumn{1}{c|}{0.2} & 0.5 & \multicolumn{1}{c|}{0.07} & \multicolumn{1}{c|}{0.2} & 0.5  \\ \hline
SHREC’19 & 6.8 & \multicolumn{1}{c|}{\textbf{4.8}} & 5.4 & \multicolumn{1}{c|}{5.4} & \multicolumn{1}{c|}{6.4} & 6.5 & \multicolumn{1}{c|}{6.2}  & \multicolumn{1}{c|}{6.5} & 6.5  \\ \hline
SCAPE-PC & 8.2 & \multicolumn{1}{c|}{\textbf{6.6}} & 7.6 & \multicolumn{1}{c|}{7.0} & \multicolumn{1}{c|}{8.7} & 8.8 & \multicolumn{1}{c|}{9.8}  & \multicolumn{1}{c|}{9.6} & 10.1 \\ \hline
\end{tabular}%
}
\caption{Ablation study on number of nodes and radius. Radius is expressed as fraction of shape diameter. Errors on two benchmarks are reported in cm.}
\label{tab:NodeRadiusAblation}
\end{table}

\subsubsection{Optimal nodes and radius}
The number of nodes and their radius are important parameters in our reduced representation. We desire a representation that is both compact and simultaneously can capture the local characteristics of deformation. Unfortunately, both of these criteria are difficult to satisfy simultaneously as it could potentially lead to a singular moment matrix (c.f Eqn 2, main draft). Therefore, we first make a choice on the compactness by letting the radius of each node be $\frac{1}{5}^{th}$ of the shape diameter. Then, from the set of candidates (Figure~\ref{fig:nodeSampling}(b)) we sample nodes until the non-singularity condition for the moment matrix is satisfied. Since the choice of radius is a hyper-parameter, we perform an ablation study by varying the radius and the number of nodes as summarized in Table~\ref{tab:NodeRadiusAblation}. We observe that by increasing the radius of each node and the number of nodes itself deters the performance. This is because larger radius impedes the \emph{locality} of the deformation by influencing distant points. On the other hand, increasing the number of nodes forces the network to learn more fine-grained details thereby showing a deteriorated performance in the setting of limited training data. Alternatively, one could first choose a fixed set of nodes and then expand their radii until the non-singularity condition is met. However, this requires a prior-knowledge on where to place the nodes and that requires manual intervention. On the other hand, our selection process is fully automatic.

\subsubsection{Positioning of nodes}
\begin{figure}[H]
\centering
\begin{subfigure}{.5\textwidth}
  \centering
  \includegraphics[width=\linewidth]{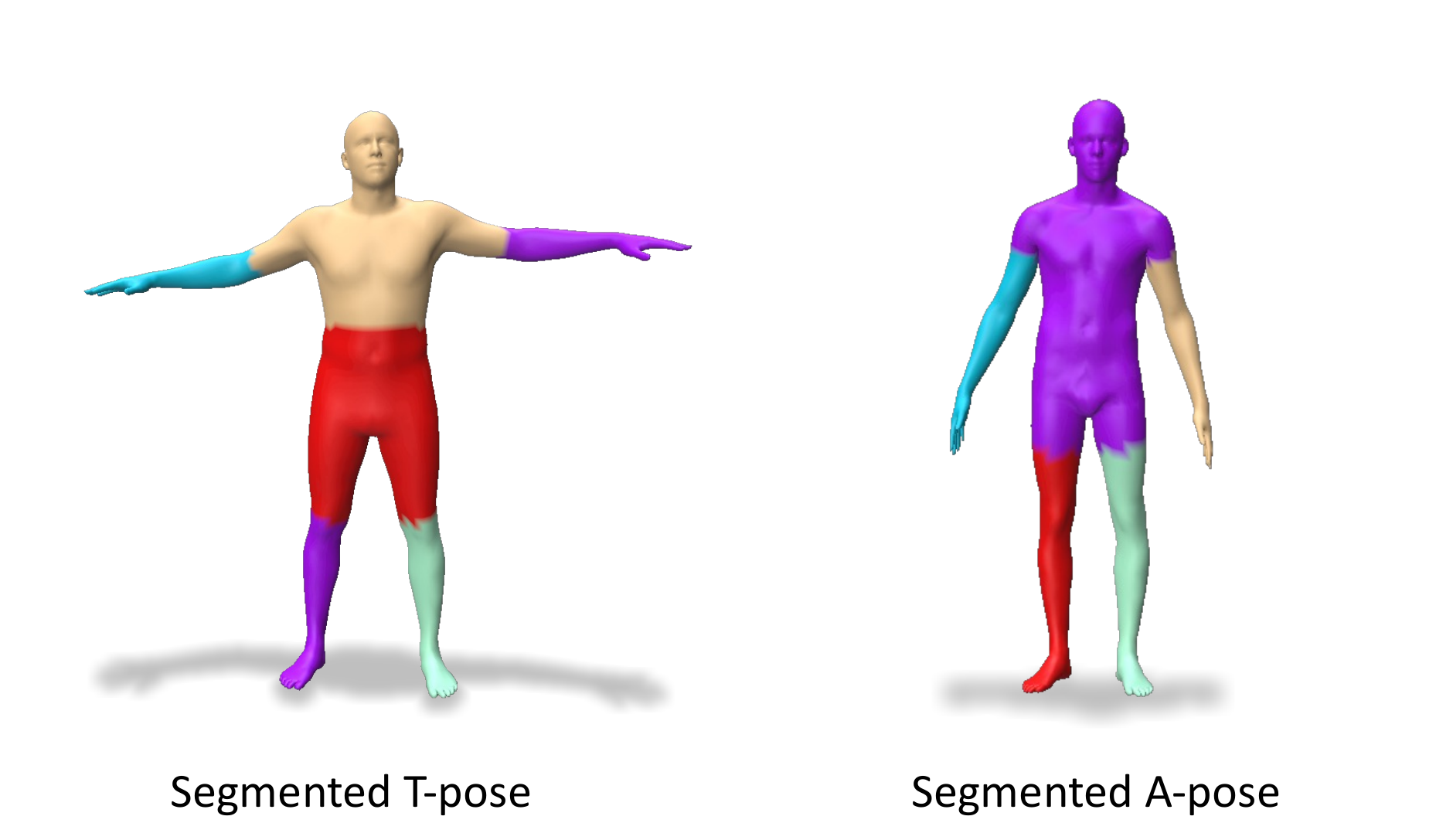}
  \caption{}
  \label{fig:sub1}
\end{subfigure}%
\begin{subfigure}{.5\textwidth}
  \centering
  \includegraphics[width=\linewidth]{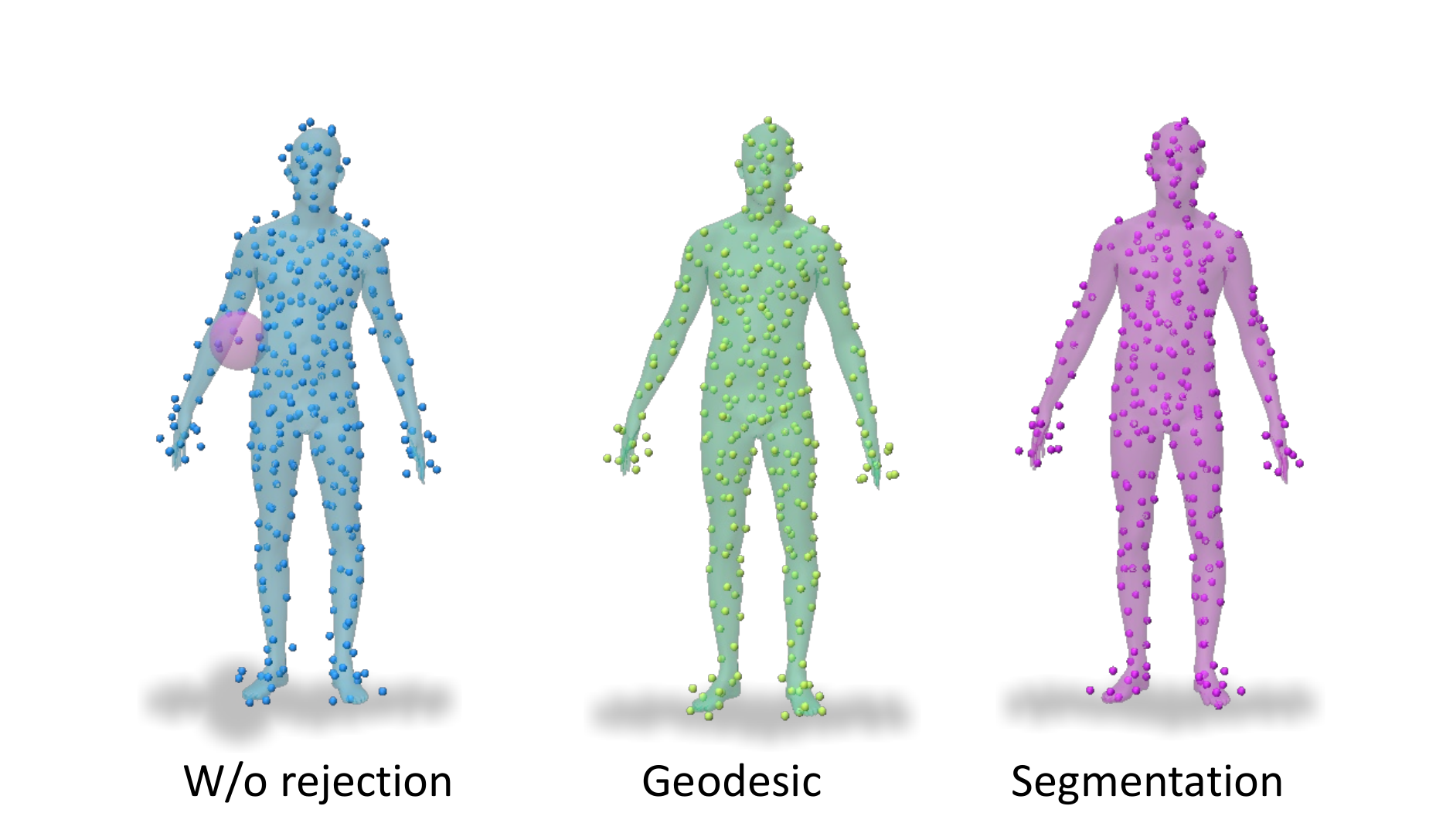}
  \caption{}
  \label{fig:sub2}
\end{subfigure}
\caption{(a) denotes pose invariant segmentation~\cite{RustamovSeg} done over T-pose and A-pose template respectively. (b) Denotes various ways of sampling nodes over A-pose template shape. Uniform sampling leads to nodes influencing points that are far in geodesic sense (highlighted in purple). This artifact is avoided by using geodesic distance on template mesh or segmentation information. Underlying surface is rendered for visualization purpose only.}
\label{fig:segSampling}
\end{figure}

Our motivation behind segmenting the template at the time of node initialization is to prevent a node from influencing the deformation field over sets of points on the template that are far from each other in a geodesic sense as this could lead to non-local deformation behavior. Since our template is from the SURREAL~\cite{varol17_surreal} dataset, we leveraged the SMPL segmentation in our main draft. In this section, we explore three alternative sampling strategies. First, we segment the template by performing K-means over the first 4 eigenvectors of the Laplace-Beltrami Operator of the template mesh as shown in Figure~\ref{fig:segSampling}. This is a well-known segmentation technique in Computer Graphics introduced by Rustamov~\cite{RustamovSeg}. This technique is unsupervised in the sense it assumes no knowledge of mesh vertex ordering. Second, we reject nodes which can influence a pair of points between which geodesic distance is larger than $20\%$ of shape diameter. Our final baseline is a simple Farthest Point Sampling (FPS) over the dense point cloud sampled over template mesh without any rejection. We compare different sampling techniques in Table~\ref{tab:samplingStrategy} over two template meshes, namely in A-pose and T-pose respectively. We observe that our proposed sampling strategy is more effective for a template in A-pose while showing marginal improvement over a straightforward sampling for a template in T-pose. This is because the likelihood of a node to influence geodesically farther (or semantically different) points is significantly higher in A-pose than in T-pose. We illustrate an example in Figure~\ref{fig:segSampling} where a node from uniform sampling is shown to influence the deformation field at both arms and torso. We note, however, that the approach based on unsupervised segmentation, performs similarly to our strategy and does not require any prior semantic information.

\begin{table}[H]
\centering
\resizebox{\textwidth}{!}{%
\begin{tabular}{@{}|c|cccc|cccc|@{}}
\toprule
Pose of Template &
  \multicolumn{4}{c|}{T-Pose} &
  \multicolumn{4}{c|}{A-Pose} \\ \midrule
Sampling Strategy &
  \multicolumn{1}{c|}{Uniform} &
  \multicolumn{1}{c|}{Geodesic} &
  \multicolumn{1}{c|}{Segmentation SMPL} &
  \multicolumn{1}{l|}{Segmentation \cite{RustamovSeg}} &
  \multicolumn{1}{c|}{Uniform} &
  \multicolumn{1}{c|}{Geodesic} &
  \multicolumn{1}{c|}{Segmentation SMPL} &
  \multicolumn{1}{l|}{Segmentation \cite{RustamovSeg}} \\ \midrule
SHREC’19 &
  \multicolumn{1}{c|}{5.1} &
  \multicolumn{1}{c|}{5.0} &
  \multicolumn{1}{c|}{\textbf{4.8}} &
  \textbf{4.8} &
  \multicolumn{1}{c|}{5.9} &
  \multicolumn{1}{c|}{5.2} &
  \multicolumn{1}{c|}{\textbf{5.1}} &
  \textbf{5.1} \\ \midrule
SCAPE &
  \multicolumn{1}{c|}{6.9} &
  \multicolumn{1}{c|}{6.9} &
  \multicolumn{1}{c|}{\textbf{6.6}} &
  \textbf{6.6} &
  \multicolumn{1}{c|}{9.5} &
  \multicolumn{1}{c|}{7.4} &
  \multicolumn{1}{c|}{7.4} &
  \textbf{7.3} \\ \bottomrule
\end{tabular}%
}
\caption{Comparison of different sampling strategy for initializing nodes across two template poses. All reported errors are in cm. }
\label{tab:samplingStrategy}
\end{table}

\subsubsection{Choice of Template}
\begin{figure}[h]
\centering
\includegraphics[width=\textwidth]{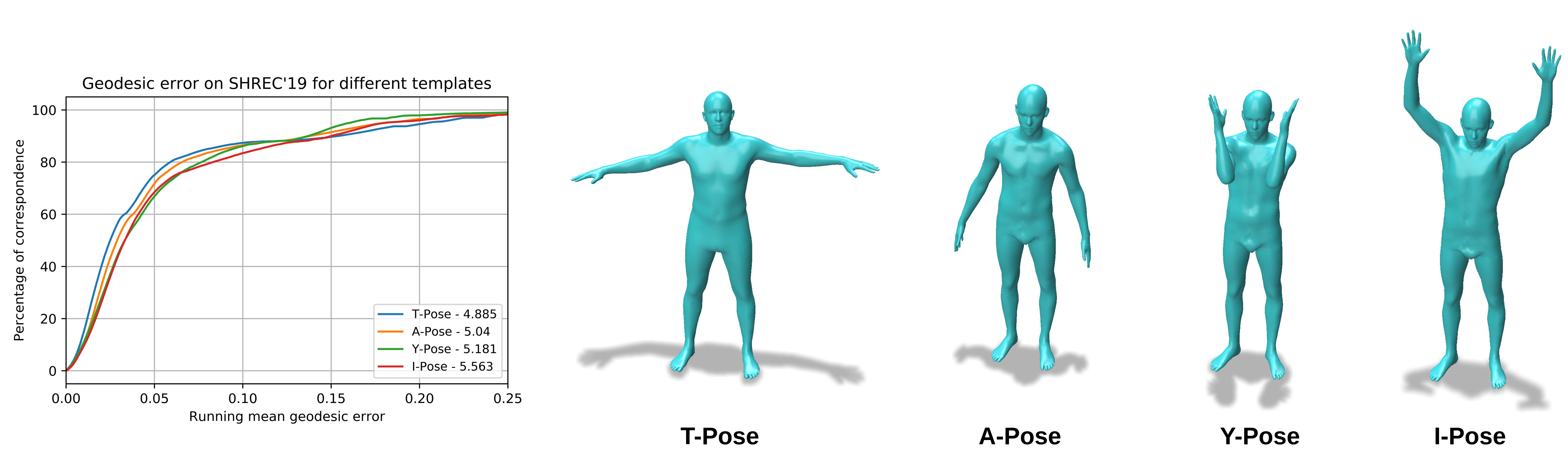}
\caption{Quantitative correspondence result on SHREC'19~\cite{Melzi2019} (left) using different templates (right).}
\label{fig:TempAblation}
\end{figure}

We analyze our choice of template by comparing with three alternatives which have different pose and style as shown in Figure~\ref{fig:TempAblation}. We compare shape-correspondence accuracy on the SHREC'19~\cite{Melzi2019} benchmark. Template in A-pose and T-pose showed comparable performance while it mildly deteriorated when using an I pose template. 

\subsection{Effect of supervision}
\label{sec:supEffect}

\subsubsection{Optimal training shapes}
\begin{figure}[h]
\centering
\includegraphics[width=\textwidth]{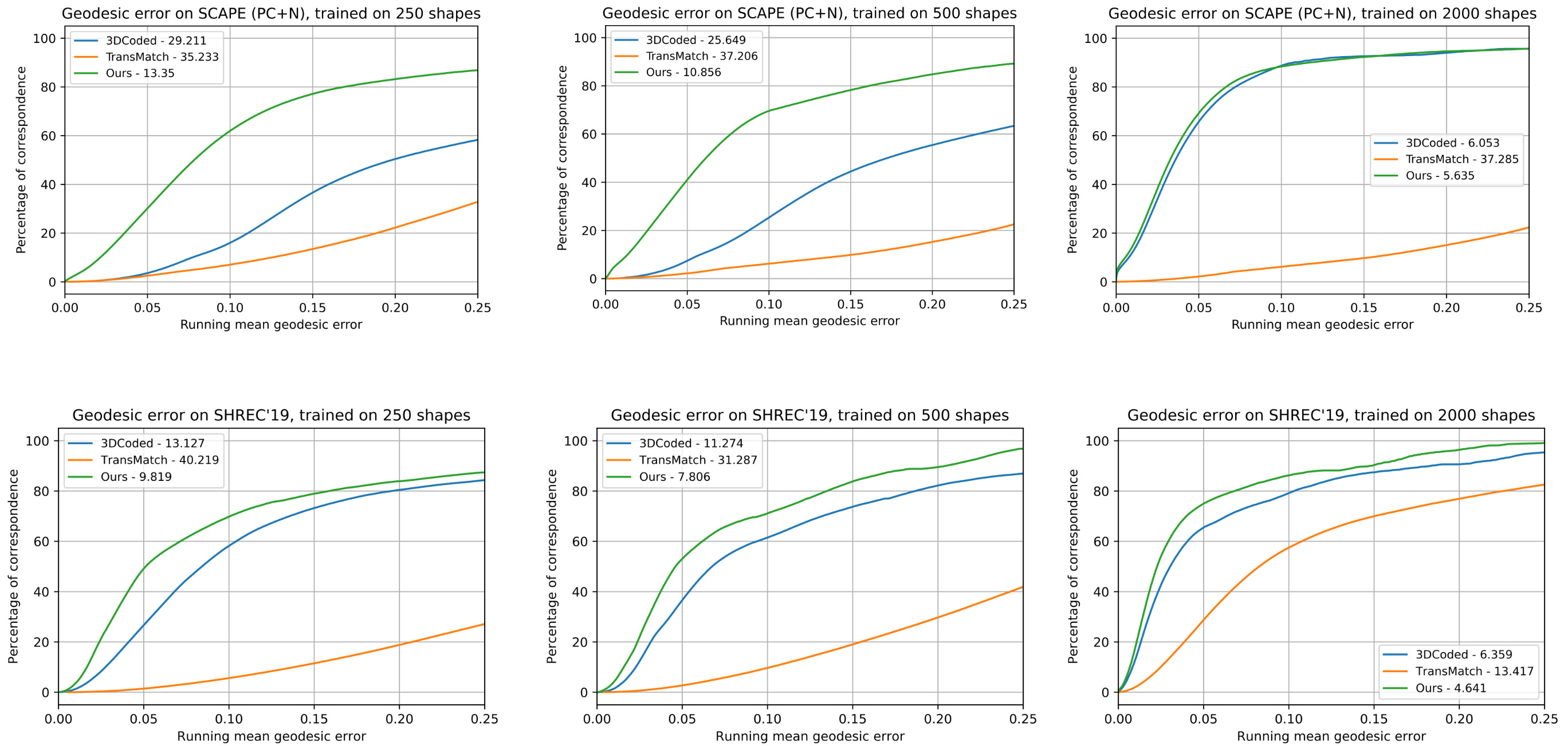}
\caption{Quantitative correspondence accuracy with varying number of shapes in the training set. Our approach shows a significant improvement over the baseline particularly when there is paucity of data.}
\label{fig:TrEffort}
\end{figure}

We simultaneously decrease and increase the amount of training data to analyze the data-dependence of various supervised deformation methods. To that end, we train our method and baselines on 250, 500, 2000 training shapes sampled at random from SURREAL dataset~\cite{varol17_surreal}. We re-train the deformation baselines TransMatch~\cite{TransMatch} and 3D-CODED~\cite{groueix20183d} on the same dataset with appropriate parameters for a fair comparison. Figure~\ref{fig:TrEffort} summarizes our comparison. Our approach shows significant improvement over 3D-CODED~\cite{groueix20183d} when trained on 250 and 500 shapes across both SCAPE (PC+N) and SHREC'19 benchmarks respectively. TransMatch~\cite{TransMatch} on the other hand fails to achieve reasonable correspondence due to the strong reliance of attention mechanism on large collection of training data.

\subsubsection{Optimal corresponding points}
\begin{figure}[h]
\centering
\includegraphics[width=0.8\textwidth]{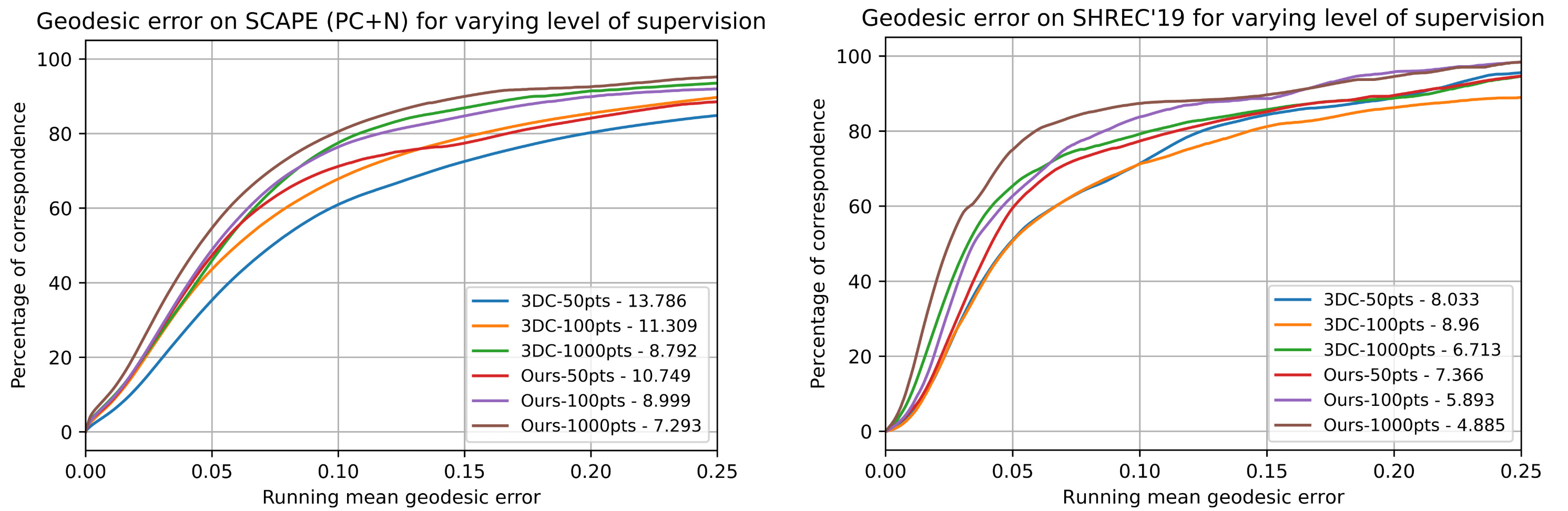}
\caption{Quantitative comparison between our approach and 3D-Coded~\cite{groueix20183d} with varying level of supervision used}
\label{fig:TrKpts}
\end{figure}

We analyze the need for supervision by comparing with 3D-CODED by varying the number of points used for supervision (c.f Equation 7 main paper). We use 50, 100 and 1000 points for supervision and compare on SCAPE (PC+N) and SHREC'19. For a fair comparison, we re-train 3D-CODED on same training shape as our method. Results are summarized in Figure~\ref{fig:TrKpts}. It is remarkable to note that our method shows improvement over the baseline with one-tenth of supervision.

\subsection{Unsupervised Implicit Shape Correspondence}
\label{sec:unusupImplicit}
\begin{figure}[h!]
\centering
\includegraphics[width=0.8\textwidth]{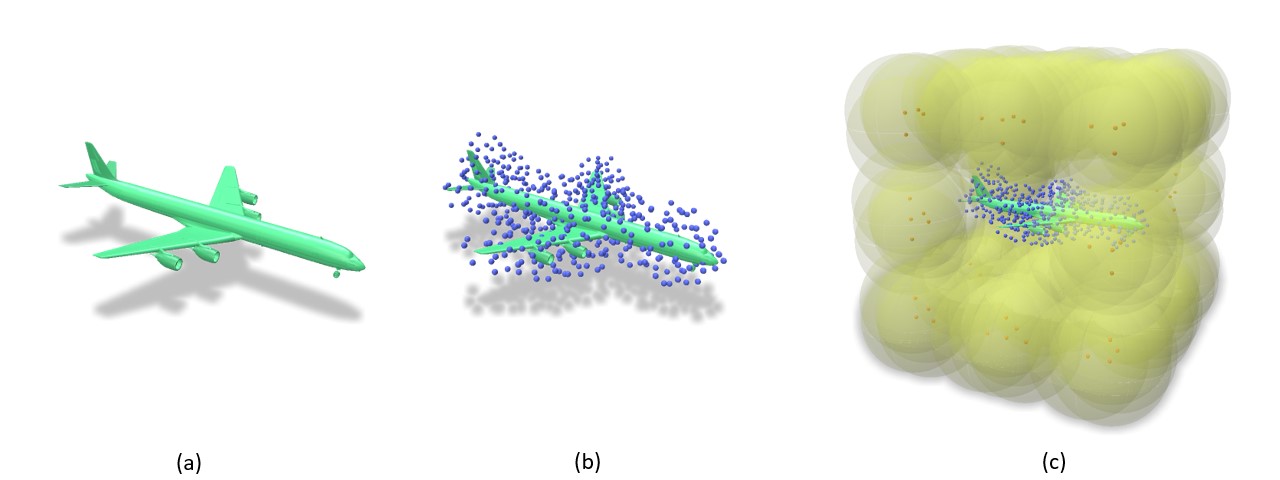}
\caption{Node sampling strategy for modelling deformation between implicit fields. (a) Given a template mesh, (b) we sample nodes near the surface. (c) Then, we define ``auxiliary'' nodes along the vertices and edge of the unit cube. Auxiliary nodes are shown in red and its spatial influence in yellow (transparent). }
\label{fig:etherQual}
\end{figure} 

 Complimentary to discussions in Section 5.3 of our main paper, we demonstrate that our proposed representation is data-efficient in establishing correspondence between learnt implicit surfaces of real-world objects from ShapeNet~\cite{shapenet2015} dataset. For this task, we adapt the existing implicit shape correspondence work DIF-Net~\cite{deng2021deformed} and replace their point-wise ``Deform-Net'' with our reduced representation. This replacement however is not straightforward as a \emph{volumetric} deformation requires the deformation field to be continuous and defined in $\mathbb{R}^3$ while our compact representation restricts to a sub-region $\Omega \subset \mathbb{R}^3$. To overcome this, we scale all shapes to fit a unit cube and place \emph{auxiliary nodes} at each vertex of the cube. These auxiliary nodes have a larger radius, thereby covering the entire region as shown in the Figure~\ref{fig:etherQual}. Please recall that since the deformation field at a point varies inversely by its distance from neighbouring nodes, the influence of auxiliary nodes are minimal near the surface of the shape. As a result, auxiliary nodes act as a regularizer for points that are far from the surface.
 
\begin{figure}[h!]
\centering
\includegraphics[width=\textwidth]{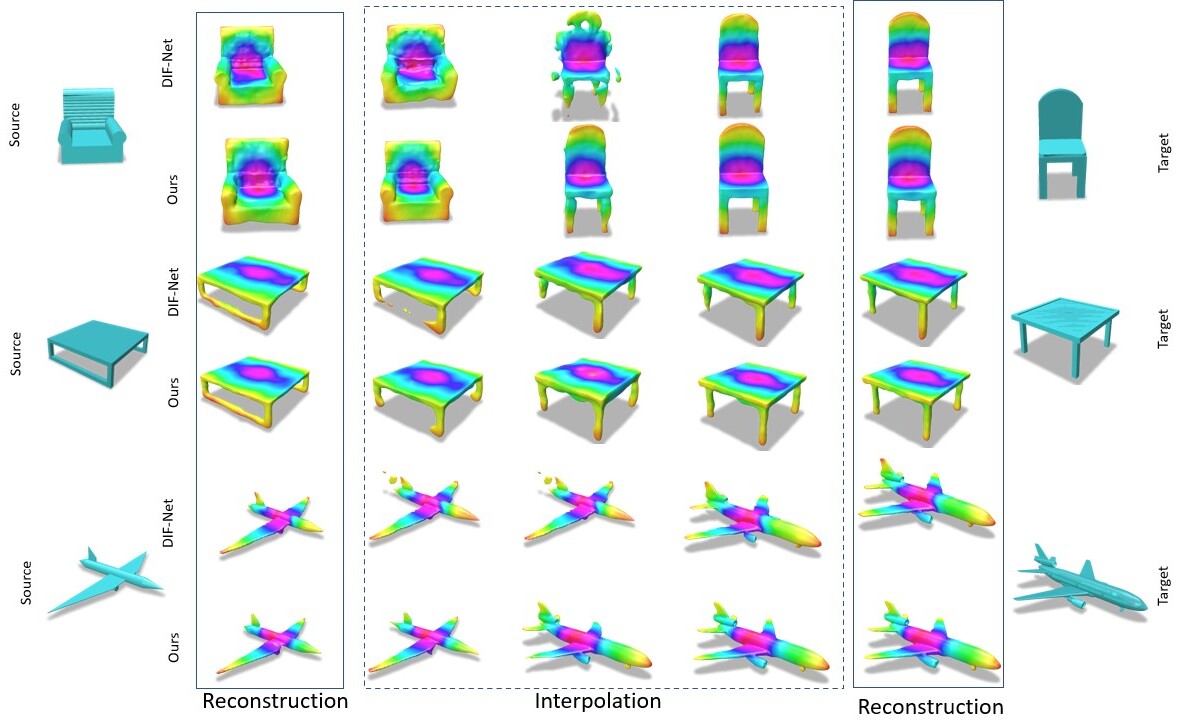}
\caption{Qualitative comparison for implicit shape correspondence across three object categories. Individual columns in interpolation corresponds to samples at same time-steps across rows. Our well-structured latent space helps to avoid implausible reconstructions.  }
\label{fig:interpQual}
\end{figure}

 We consider three categories namely chair, table and plane. Since there is no dense ground-truth correspondence annotation between them, we show qualitative interpolation results. For a fair comparison, we train both DIF-Net and our method over same shapes, consisting of 500 random samples from each category. We summarize our qualitative results in Figure~\ref{fig:interpQual}. Our method produces \emph{plausible} interpolation sequences owing to the latent constraint $\mathcal{L}_Z$ that regularizes the deformation field corresponding to intermediate shapes. In addition, our deformation field is smooth as shown by the color-map.

\subsection{Real-world Data}
\label{sec:realWorlddata}
Finally, we demonstrate the generalization of our approach in solving correspondence between real-world data over two datasets, namely, point cloud from RGB-D scans of humans and meshes of hearts. \\

\subsubsection{CMU Panoptic Dataset}
\begin{figure}[h!]
\centering
\includegraphics[width=\textwidth]{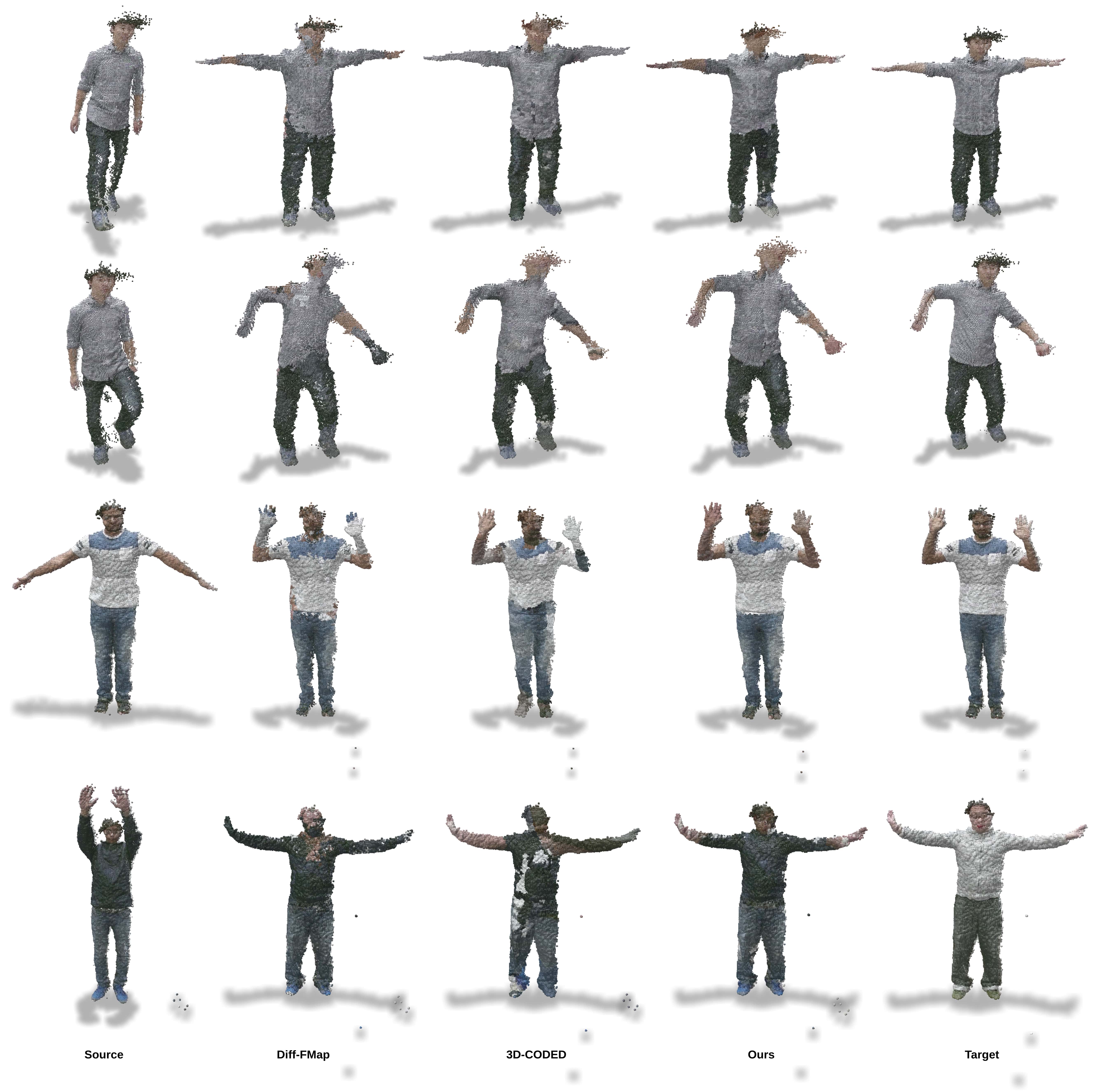}
\caption{Qualitative comparison on real-world data from the CMU Panoptic dataset with noise and outlier points. First three rows show texture transfer between same subjects while the last row is an example of inter-subject transfer.}
\label{fig:realworldData}
\end{figure}

We show qualitative results of texture transfer between point clouds obtained from \emph{Kinect RGB-D sensor}, from the CMU Panoptic dataset~\cite{Joo_2017_TPAMI}. Our results are shown in Figure~\ref{fig:realworldData}. Last row shows a particularly interesting example of texture transfer between different subjects. Our approach is the only method that provides reasonable correspondence. 

\subsubsection{Human Heart Meshes}
\begin{figure}[h!]
\centering
\includegraphics[width=\textwidth]{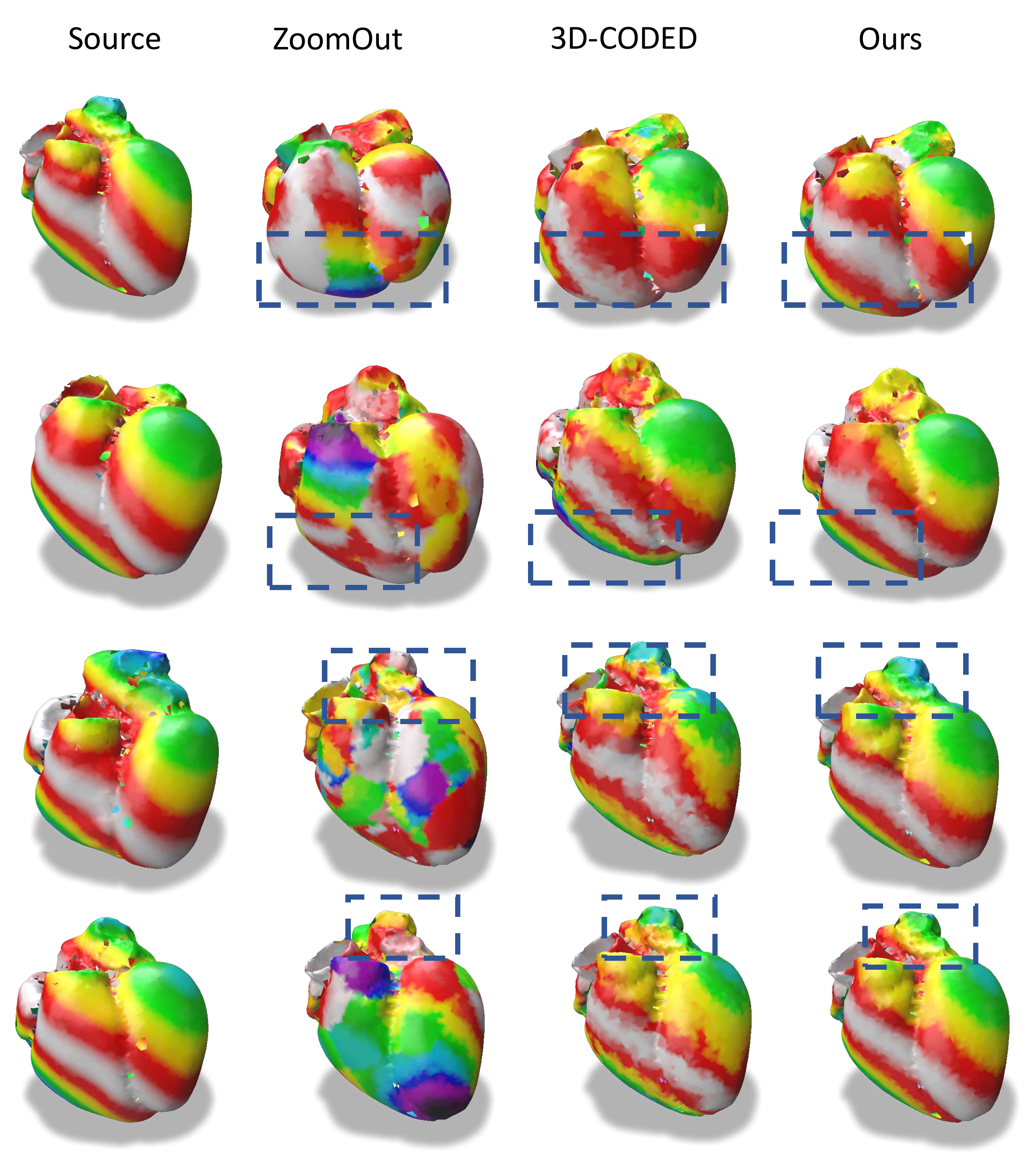}
\caption{Qualitative comparison of colour-transfer between human hearts of different subjects obtained from ~\cite{Strocchi2020}. Dotted box highlights the efficacy of our approach in computing smooth and consistent map.}
\label{fig:heartScan}
\end{figure}

In order to demonstrate the applicability of our novel deformation-field representation beyond articulated non-rigid shapes, we consider the publicly available virtual cohort of four-chamber heart meshes dataset~\cite{Strocchi2020}. This dataset is generated from twenty-four heart failure (HF) patients, starting from CT scan images of the heart. These images are then segmented to distinguish the four chambers and a tetrahedral mesh is constructed from the resulting segments. In our experiments, we use the outer boundary surface of the mesh after Quadratic Edge Collapse Decimation~\cite{QECDecimation}. Qualitative results are shown in Figure~\ref{fig:heartScan}. We compare our approach with ZoomOut~\cite{Ren2018}, an axiomatic shape correspondence technique and our closest deformation-based baseline, 3D-CODED~\cite{groueix2018b}. Our approach extracts a more consistent correspondence in comparison to the two baselines. We attribute this to our regularization that facilitates smoothness and learning over a small-scale dataset.

\bibliographystyle{unsrt}
\bibliography{shapemaps}

\end{document}